\documentclass[hidelinks,onefignum,onetabnum]{siamart251216}

\usepackage{lipsum}
\usepackage{amsfonts}
\usepackage{graphicx}
\usepackage{epstopdf}
\usepackage{algorithmic}
\usepackage{booktabs}
\usepackage{adjustbox}
\usepackage{multirow} 
\usepackage{hyperref}

\ifpdf
  \DeclareGraphicsExtensions{.eps,.pdf,.png,.jpg}
\else
  \DeclareGraphicsExtensions{.eps}
\fi

\newsiamremark{remark}{Remark}
\newsiamremark{hypothesis}{Hypothesis}
\crefname{hypothesis}{Hypothesis}{Hypotheses}
\newsiamthm{claim}{Claim}
\newsiamremark{fact}{Fact}
\crefname{fact}{Fact}{Facts}

\headers{Robust Student's t Likelihood for BNNs}{P. Hsia, L. H. Heyen, A. Weyrauch, M. Goetz, A. Streit, S. Krumscheid, C. Debus}

\title{Rethinking Likelihood distributions: Student's t Likelihood Boosts Bayesian Neural Network Performance\thanks{Submitted to the editors July 28, 2026.
\funding{This work is supported by the Helmholtz Association Initiative and Networking Fund under the Helmholtz AI platform grant.}}}

\author{Pei-Hsuan Hsia\thanks{Scientific Computing Center, Karlsruhe Institute of Technology.}
\and Lars H. Heyen\footnotemark[2]
\and Arvid Weyrauch\footnotemark[2]
\and Markus Goetz\thanks{Scientific Computing Center, Karlsruhe Institute of Technology; and Helmholtz AI.}
\and Achim Streit\footnotemark[2]
\and Sebastian Krumscheid\footnotemark[2]
\and Charlotte Debus\thanks{Scientific Computing Center, Karlsruhe Institute of Technology (\email{charlotte.debus@kit.edu}).}}

\usepackage{amsopn}

\def\weightsymb{W}
\def\weightmean{\mu}
\def\weightstd{\sigma}

\def\vardist{q}
\def\varparams{\lambda}
\def\prior{p}

\def\kl{\mathrm{KL}}

\def\trainsamples{\mathbf{X}}
\def\trainlabels{\mathbf{Y}}

\newcommand{\expv}[2]{\mathbb{E}_{#2}\left[ #1 \right]}
\newcommand{\given}{\, | \,}

\ifpdf
\hypersetup{
  pdftitle={Rethinking Likelihood distributions: Student's t Likelihood Boosts Bayesian Neural Network Performance},
  pdfauthor={Pei-Hsuan~Hsia, Lars~H.~Heyen, Arvid~Weyrauch, Markus~Goetz, Achim~Streit, Sebastian~Krumscheid, Charlotte~Debus}
}
\fi

\begin{document}

\maketitle

\begin{abstract}
    In Bayesian neural networks (BNNs), variational inference is a widely adopted framework for modeling uncertainty in a distributional way, with the evidence lower bound (ELBO) serving as the standard objective function. 
    Several distributions contribute to the ELBO loss, such as the prior, approximated posterior, and likelihood distribution. 
    Typically, these distributions are all approximated by a Gaussian distribution, since it is easy to compute, allows for reparameterized gradients, and provides a closed-form loss for training.
    However, several works have highlighted that this assumption may not generally hold, posing the risk of model misspecification. 
    Alternative distributions have been proposed for the prior specifically, while the effect of distribution choice on the likelihood distribution remains unexplored. 
    In this work, our aim is to close this gap by investigating whether alternative assumptions for the likelihood distribution can outperform the commonly used Gaussian.
    We compare several likelihood distribution assumptions, such as skewed or heavy-tailed, across regression tasks on both artificial and real-world datasets using standard multilayer perceptrons (MLPs). 
    Our findings demonstrate that Student’s t yields better predictive performance than a Gaussian likelihood distribution, independent of the data distribution and MLP architecture (depth and width). 
    In some cases, Student's t can also lead to shorter training times, while still being easy to implement.
\end{abstract}

\begin{keywords}
Uncertainty quantification, Bayesian neural networks, Likelihood, Student's t
\end{keywords}

\section{Introduction}
Uncertainty quantification is a crucial aspect of machine learning, particularly when developing models that can be trusted in real-world applications, for example, autonomous driving~\cite{kendall2017uncertainties}, energy systems~\cite{hong2016probabilistic}, seismology~\cite{armstrong2023deep}, and finance~\cite{nagl2022quantifying}.
The AI Act of the European Union specifically requires high-risk AI systems to demonstrate appropriate levels of accuracy, robustness, and cybersecurity, reinforcing the need for documented uncertainty estimates~\cite{EUAIAct2024}.
Neural networks, which have become a fundamental component of numerous successful machine learning applications, traditionally provide only point predictions without offering any information on the uncertainty associated with those predictions.
Without uncertainty estimation, reliability cannot be quantified, hindering risk-aware decision-making in critical infrastructures, e.g., renewable energy planning.
Bayesian neural networks (BNNs)~\cite{neal1992bayesian,mackay1995bayesian} offer a powerful solution to uncertainty quantification in neural networks by incorporating uncertainty into the model through probabilistic inference.
Instead of optimizing weights with fixed values, BNNs attempt to obtain the weight posterior distribution given the training set of inputs and labels.
This also transforms point predictions into likelihood distributions.
Among different BNN approaches, variational inference (VI)~\cite{blei2017variational,hoffman2013stochastic} avoids the intractable integration over weights required by exact Bayesian inference via a tractable variational approximation of the posterior.

As the dominant choice of loss function, the Evidence Lower BOund (ELBO)~\cite{blei2017variational} holds a central role in the training of a BNN with VI. 
Three key distributions feed into the ELBO: The prior, the approximate posterior and the likelihood distribution. While the prior represents background knowledge of the parameters, the approximate posterior represents the updated belief about these parameters after observing the data. 
The likelihood distribution represents the expected distribution of the outputs for new inputs under the model, reflecting uncertainty in the predictions.
To calculate the ELBO during BNN training, we must choose a type or distribution to represent the prior, the approximation of the posterior, and the likelihood distribution.
A common choice is the Gaussian~\cite{blundell2015weight,khan2018fast,kingma2015variational}, since it is easy to compute and generalizes the likelihood term akin to the mean squared error.
However, some works have pointed out that model misspecification, including incorrect assumptions of these three distributions, may be a problem that limits the performance of BNNs~\cite{izmailov2021bayesian,dusenberry2020efficient}. 
Hence, it is crucial to test the validity of the Gaussian assumption and, if necessary, to find a more suitable alternative distribution.
During training, the model is supposed to learn the relationship between the input and the output.
Given an input and a perfectly trained model, the likelihood distribution thus should capture the distributional form of the noise of the data.
This raises the question of whether matching the likelihood distribution to the data noise provides the best solution, making the choice of likelihood distribution data dependent, or whether a data-independent distribution can be found that universally outperforms the Gaussian.

Much of the BNN literature has focused on investigating alternative prior distributions~\cite{hernandez-lobato_probabilistic_2015, ghosh2019model, carvalho_handling_nodate} or efficient parameterizations of the approximate posterior to manage computational loads~\cite{dusenberry2020efficient}. While works on approaches other than VI, e.g. for Bayesian linear regression~\cite{shah2014student}, Gaussian processes~\cite{jylanki2011robust,lee2021scale}, and spatiotemporal prediction~\cite{saad2024scalable}, have studied the potential of non-Gaussian likelihood distributions, a systematic analysis of the Gaussian assumption for the likelihood distribution specifically within the VI framework is still lacking. 

In this study, we aim to close this gap and explicitly test the hypothesis of whether the likelihood distribution should mirror the data noise by conducting controlled experiments with data of known noise distribution---and find that it is not necessary for optimal performance. 
In fact, we find that a heavy-tailed Student’s $t$ distribution generally performs best across datasets, which better represent the dataset distribution.
We validate these findings on two real-world tasks, and analyze convergence time and stability. 
In general, the choice of the likelihood distribution has a notable effect on predictive accuracy in VI, and we find Student’s $t$ to be a robust default choice.

\section{Related Work}
\label{sec:related_work}

Several previous studies have investigated specific aspects of the ELBO loss commonly used in BNNs.
Within ELBO, a notable problem discussed in the literature is the potential misspecification of the prior distribution. 
For instance, recent research has highlighted that an inappropriate prior choice can negatively impact the quality of uncertainty estimates and predictive performance~\cite{fortuin2021bayesian}. 
Addressing this issue, researchers have proposed more flexible or adaptive priors to mitigate prior misspecification, leading to improved model robustness and predictive uncertainty calibration~\cite{izmailov2021bayesian,dusenberry2020efficient}. 
However, prior misspecification remains challenging, particularly with regard to computational complexity, leaving room for further improvement.

Model misspecification is a larger issue that is extensively discussed in Bayesian inference literature~\cite{fortuin2021bayesian, knoblauch2019generalized}. 
This problem arises when the true data-generating distribution differs significantly from the assumed likelihood distribution or the prior. 
Proposed solutions typically involve introducing more flexible modeling assumptions, such as mixture models, heavy-tailed distributions, or hierarchical Bayesian approaches~\cite{gelman2017prior,huggins2019robust}.
Although these methods have shown success in reducing misspecification effects, their complexity and computational demands can limit widespread adoption. 
As a result, developing computationally efficient yet robust methods to handle model misspecification remains an open challenge.

The use of Student’s $t$-distribution in Bayesian predictive modeling has been investigated in various statistical contexts outside BNNs.
For instance, studies in Bayesian linear regression~\cite{shah2014student}, Gaussian processes~\cite{jylanki2011robust,lee2021scale}, dynamic neural regression models \cite{briegel2000dynamic} and Bayesian neural fields for spatiotemporal prediction \cite{saad2024scalable} have demonstrated the advantages of using Student’s $t$ likelihood distributions, particularly in the presence of noisy and heavy-tailed data.
Alternative likelihood distributions have been mentioned in traditional BNN literature using Markov Chain Monte Carlo methods~\cite{lampinen2001bayesian}, and recent studies have proposed modifying the ELBO through robust divergences~\cite{futami2018variational}. 
However, these works do not provide a systematic analysis of likelihood distribution selection specifically for BNNs trained via VI. 
We aim to fill this gap in the literature with a systematic study of the effect of the choice of likelihood distribution on BNN accuracy.

\section{Variational Inference}\label{sec:variational-inference}

Although the Bayesian framework in principle allows us to infer the posterior distribution $p(\weightsymb\given\trainsamples, \trainlabels)$ over weights $\weightsymb$, the exact computations are intractable for neural networks. 
According to Bayes’ rule, the posterior is
\begin{equation}
    p(\weightsymb \mid \trainsamples,\trainlabels) \propto p(\trainlabels \mid \trainsamples,\weightsymb)\,\prior(\weightsymb),
\end{equation}
where $\trainsamples$ is the input of the training set, $\trainlabels$ is the labels, and $p(W)$ is the prior.
The quantity needed for prediction (for a new input \(x'\) after training) is the output distribution obtained by integrating over the posterior:
\begin{equation}
p(y' \mid x',\trainsamples,\trainlabels)
= \int p(y' \mid x',\weightsymb)\, p(\weightsymb \mid \trainsamples,\trainlabels)\, d\weightsymb .
\label{eq:predictive_marginal}
\end{equation}
For modern networks, the posterior $p(\weightsymb \given \trainsamples,\trainlabels)$ lacks a closed form, and the predictive integral in \Cref{eq:predictive_marginal} is high-dimensional (the number of trainable parameters) and analytically intractable, making exact computation infeasible.

An established method to overcome the intractable computation of the posterior $p(\weightsymb \given \trainsamples, \trainlabels)$ is Variational Inference (VI)~\cite{blei2017variational}.
This method approximates the true posterior $p(\weightsymb\given\trainsamples, \trainlabels)$ by optimizing the parameters $\varparams$ of a variational, i.e., parameterized distribution $\vardist(\weightsymb\given\varparams)$.
This approximated posterior distribution is the current best estimate of the true posterior during training.
Typically, VI is performed under the mean-field approximation, which assumes that all correlations between the model parameters are zero~\cite{zhang2018VI_advances}.
Although it is well known that this approximation is not entirely accurate~\cite{fortuin2021bayesian}, the resulting reduction in model complexity---especially the number of trainable parameters---is necessary to make larger BNNs realistically trainable.

In this paper, we consider models that use an uncorrelated multivariate Gaussian distribution as an approximated posterior distribution.
The approximated posterior distribution is parameterized by the means $\weightmean$ and standard deviations $\weightstd$ of the weights.
Because we learn a distribution over weights rather than a single point estimate, the training objective must also be distributional. 
VI optimizes \(\vardist\) to be close to the true posterior by minimizing the Kullback–Leibler divergence ($\kl$) \cite{kullback1951information} 
\begin{equation}
\begin{split}
    \kl[\vardist(\weightsymb\mid\varparams)\,\|\,p(\weightsymb\mid\trainsamples,\trainlabels)].
\end{split}
\end{equation}
Rearranging this quantity and removing terms independent of the weights yields the Evidence Lower BOund (ELBO)~\cite{neal1998view},
\begin{equation}
\begin{split}
    \mathrm{ELBO} = \,&\expv{\log p(\trainlabels\given\trainsamples, \weightsymb)}{\weightsymb \sim \vardist}\\
    - &\mathrm{KL}[\vardist(\weightsymb\given\varparams)||\prior(\weightsymb)]~,
    \label{eq:elbo}
\end{split}
\end{equation}
where the first term
is the expected value of the log-likelihood of the labels $\trainlabels$ with respect to $\weightsymb$ sampled from $\vardist$ and given model inputs $\trainsamples$, and 
the second term is the $\kl$ divergence between the approximated posterior distribution and the prior.
In practice, we optimize the ELBO with respect to the variational parameters $\varparams$ using Bayes-by-Backprop~\cite{blundell2015weight}.

\section{Likelihood distribution and Research Question}

We refer to the conditional probability density of $\trainlabels$ given $\trainsamples$ and $\weightsymb$, $p(\trainlabels \given \trainsamples,\weightsymb)$, as the likelihood distribution.
The model's agreement with the data is assessed by the log-likelihood of the labels under this distribution, i.e., evaluating \(\log p(\trainlabels \given \trainsamples,\weightsymb)\) with the ground truth \(\trainlabels\). 
In VI, this appears as \(\mathbb{E}_{\weightsymb\sim \vardist}\!\left[\log p(\trainlabels \given \trainsamples,\weightsymb)\right]\) in the ELBO. 
The distribution’s mean is used as the point prediction, and its distributional form characterizes the predictive intervals (uncertainty). 
For deterministic neural networks, optimization of network weights reduces to minimizing a standard loss such as mean squared error (MSE), which corresponds to the negative log-likelihood of a Gaussian observation model with fixed variance.

In VI, the parameters of the likelihood distribution \(p(\trainlabels \given \trainsamples,\weightsymb)\) are derived empirically from the stochastic forward passes of the BNN. 
The predictive mean $\hat{\mu}$ and standard deviation $\hat{\sigma}$ are then computed as the first moment (sample mean) and the second central moment (sample variance) to define the location and scale of the likelihoods. 
For the skew-normal distribution, we further incorporate the third standardized moment (skewness) to adaptively adjust the shape parameter $\hat{\gamma}$ ($= \mathbb{E}[((X - \mu)/\sigma)^3]$).

These statistics are utilized to parameterize the candidate distributions as follows (exact formulas shown in Appendix~\ref{app:preddists}):
\begin{itemize} 
\item \textbf{Gaussian:} The distribution is defined by parameters $(\mu, \sigma)$, where we set $\mu=\hat{\mu}$ and $\sigma=\hat{\sigma}$.

\item \textbf{Skew-normal:} To account for potential asymmetry in the predictive uncertainty, we derive the location $\xi$, scale $\omega$, and shape $\alpha$ from the sample statistics. We first calculate the correlation index $\delta$ using a numerical approximation based on the sample skewness $\hat{\gamma}$:
\begin{equation}
    |\delta| = \sqrt{\frac{\pi}{2} \frac{|\hat{\gamma}|^{2/3}}{|\hat{\gamma}|^{2/3} + ((4-\pi)/2)^{2/3}}}.
\end{equation}
The shape parameter $\alpha$, scale $\omega$, and location $\xi$ are then reconstructed as:
\begin{equation}
    \alpha = \frac{\delta}{\sqrt{1-\delta^2}}, \quad \omega = \hat{\sigma}\sqrt{\frac{\pi}{\pi - 2\delta^2}}, \quad \xi = \hat{\mu} - \omega\delta\sqrt{\frac{2}{\pi}}.
\end{equation}

\item \textbf{Student's $t$:} We fix the degrees of freedom $\nu = 5$ across all experiments. The location parameter is set to $\hat{\mu}$, while the scale parameter $\tau$ is derived to align the $t$-distribution's variance with the sample variance:
\begin{equation}
    \tau = \hat{\sigma} \sqrt{\frac{\nu - 2}{\nu}} + \epsilon,
\end{equation}
where $\epsilon = 10^{-5}$ is a small constant added for numerical stability to prevent underflow in the log-likelihood calculation.
\end{itemize}

This empirical parameterization ensures that the likelihood directly reflects the predictive uncertainty captured by the weight samples.
Note that in our controlled datasets, the only source of uncertainty in \(\trainlabels\) is the added noise \(\varepsilon\).
Repeated measurements at the same input would scatter around the true value according to that noise distribution. 
With controlled artificial datasets, we can test alternatives to the commonly assumed Gaussian because the noise distribution is known. 
We therefore investigate whether distributions tailored to the data noise characteristics perform better than the Gaussian likelihood distribution assumption.

\section{Experimental Evaluation}\label{sec:experimental-setup}

To test our hypothesis, we suppose that if the likelihood distribution is assumed to be of the same distributional family as the output noise, it should best capture the output distribution. 
To verify this point, we train neural networks with different shapes on datasets with different noise, comparing the test accuracy and training time for different choices of the likelihood distribution.

\subsection{Compute Infrastructure}

All experiments are conducted on a single node of a supercomputing cluster, equipped with two Intel Xeon ``Ice Lake'' processor cores, four NVIDIA A100 Tensor Core GPUs, and 512 GB main memory, operating on Red Hat Enterprise Linux (RHEL) 9.4.
The code was implemented and run under Python version 3.9 and uses PyTorch~\cite{ansel2024pytorch} version 2.6.0+cu124 with CUDA version 12.4. and torch-bayesian version 0.0.1~\cite{torchbayesian}. 
Runs were performed using a single NVIDIA A100-40 GPU for each model.
Source code is available at~\url{https://anonymous.4open.science/r/predictive_distributions-FF48/}.

\subsection{Univariate Artificial Datasets for Initial Experiments}\label{sec:uniartificial}
\begin{figure}[!h]
\centerline{\includegraphics[height=2.0in]{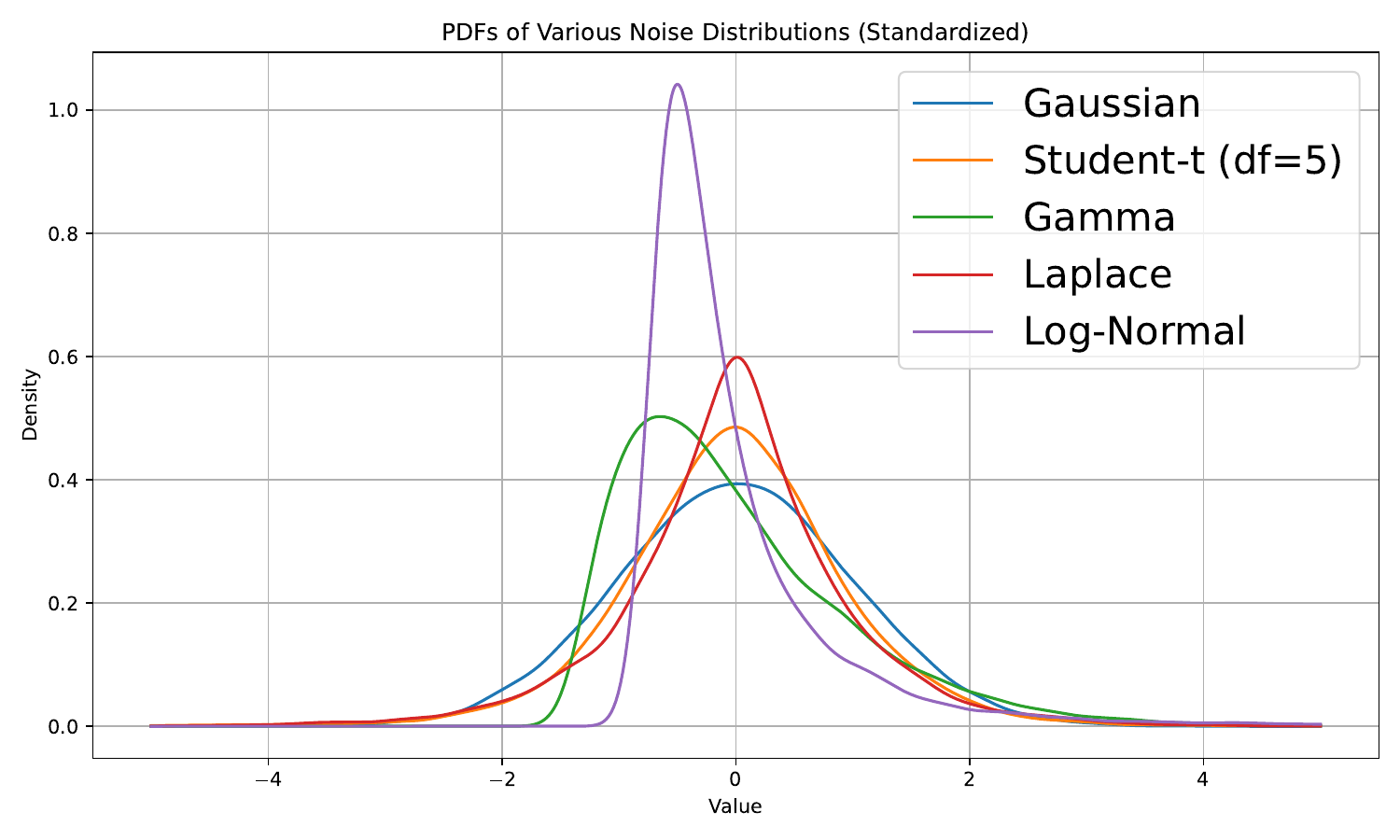}}
\caption{Standard PDFs of all noises.} \label{fig:noise_pdf}
\end{figure}

We start by generating a univariate artificial datasets, consisting of pairs $(x,y)$, with inputs $x$ uniformly sampled from $[-5,5]$ and  $y=x^3+\varepsilon$ as a function of $x$ and noise $\varepsilon$.
The noise $\varepsilon$ follows three distinct distribution families: Gaussian, Gamma, and Student’s $t$. 
These were specifically selected to cover both the Gaussian default assumption and primary deviations from Gaussian properties: specifically, asymmetry (Gamma) and heavy-tailed (Student’s $t$).
The probability density functions of all noise distributions, standardized to mean $0$ and variance $1$, are shown in \Cref{fig:noise_pdf}.

For the artificial datasets, the mean and variance of the noises are shifted and scaled to mean $0$ and standard deviation $3$.
We create datasets of three different sizes, consisting of 300, 3,000, and 30,000 data samples, respectively, to test the performance on large and small datasets.
Each dataset is split into training, validation, and testing subsets with an 8:1:1 ratio. 
Data are shuffled and min-max normalized before training.

\begin{figure}[!h]
\centerline{\includegraphics[height=2.1in]{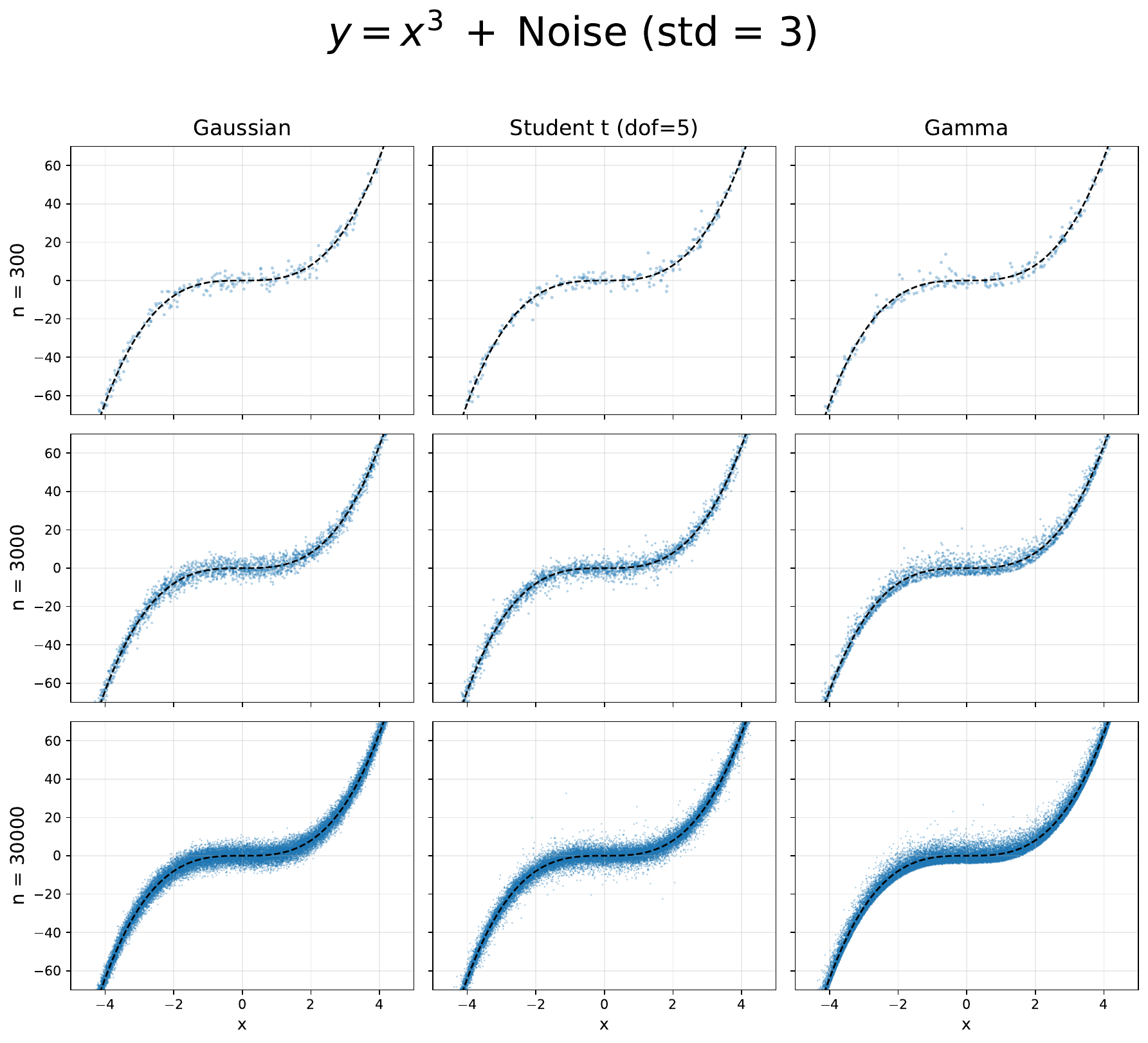}}
\caption{Scatter plots of univariate dataset.} \label{fig:scatter}
\end{figure}

\subsection{Multivariate Artificial Datasets for Scalability Analysis}\label{sec:multiartificial}

To further verify the generalizability of our findings to higher-dimensional scenarios, we extend our evaluation to a more complex artificial dataset, where the model must capture non-linear dependencies across multiple output dimensions.
We consider an input space $\mathbf{x} \in \mathbb{R}^{20}$, where each component $x_i$ is independently sampled from a uniform distribution:
\begin{equation}
    x_i \sim \mathcal{U}(-3, 3), \quad i = 1, \dots, 20.
\end{equation}

The target vector $\mathbf{y} \in \mathbb{R}^{5}$ is generated through a multivariate mapping $f: \mathbb{R}^{20} \to \mathbb{R}^{5}$ with additive noise $\boldsymbol{\varepsilon}$:
\begin{equation}
    \mathbf{y} = f(\mathbf{x}) + \boldsymbol{\varepsilon} \, ,
\end{equation}
where the ground-truth mapping $f(\mathbf{x}) = [y_1, \dots, y_5]^\top$ is defined as follows:
\begin{align}
    y_1 &= (x_1^2 + \sin(x_2 x_3)) \exp(-0.2|x_4|) \nonumber \\
    y_2 &= 10\tanh(x_5 + x_6) + 5|x_7 - x_8|^{1.5} \nonumber \\
    y_3 &= 7\cos(x_9^2 + x_{10}) + \sqrt{|x_{11}|+1} \cdot x_{12} \nonumber \\
    y_4 &= \ln(1+e^{x_{13}x_{14}}) - \ln(1+e^{x_{15}+x_{16}}) \nonumber \\
    y_5 &= \sum_{i=17}^{20} (i-16) \sin(x_i)
\end{align}

This formulation incorporates a diverse set of mathematical properties, including periodic oscillations, saturating non-linearities, and varying sensitivity across input dimensions. 
Compared to the first dataset, we expand the noise distributions $\varepsilon$ to include Laplace and Lognormal cases, providing a broader coverage of heavy-tailed and skewed deviations (see \Cref{fig:noise_pdf}). The corresponding probability density functions are shown in Appendix~\ref{app:noisedists}.

We test two signal-to-noise ratios by scaling the noise to a standard deviation of $3$ and $15$. 
To assess the scalability of the variational approximation, we again generate datasets of two different scales: $N = 10,000$ and $N = 50,000$ samples.
Following the same experimental protocol as in \Cref{sec:uniartificial}, the data is partitioned with an 8:1:1 ratio, shuffled, and min-max normalized prior to training.

\subsection{Real World Datasets}\label{sec:realdata}

We further conducted experiments on two real-world datasets to validate our findings from the controlled artificial data sets.
For one, choose a regression task using the Combined Cycle Power Plant dataset from the UCI Machine Learning repository \cite{combined_cycle_power_plant_294}, which includes 9,568 observations with four input features (temperature, ambient pressure, relative humidity, and exhaust vacuum) and one continuous output variable representing power output. 
Further, we selected a time series forecasting task based on electricity demand data for Germany from the ENTSO-E 
\cite{platform2022entso}. 
For this task, the model uses the past 8$\times$24 hours of consumption data to predict the next 2$\times$24 hours.

\subsection{Models}\label{sec:model}

We use a simple multilayer perceptron (MLP) with ReLU activation functions, and different network depths and widths. 
The model architectures are defined by their depth (number of hidden layers) and width (nodes per hidden layer). 
In the univariate scenario, we explore several configurations: 1$\times$32, 2$\times$16, 4$\times$8, and 8$\times$4.
For multivariate datasets, we employ larger architectures to accommodate higher feature complexity, specifically 2$\times$64, 2$\times$128, 4$\times$64, and 4$\times$128.
For real-world datasets, instead, we consider neural networks with different depths (1, 4, 8) and widths (4, 16, 32), resulting in nine distinct architectural configurations.
These MLPs are augmented with mean-field variational inference (hereon referenced as VIMLPs) to promote them to BNNs.
The prior and approximate posterior are assumed to be Gaussian throughout all experiments, with the prior mean set to $0$ and the prior standard deviation set to $1$.

Training of the variational parameters is performed using Bayes-by-Backprop~\cite{blundell2015weight} with the Adam optimizer~\cite{kinga2015method} with a fixed learning rate $10^{-3}$ without weight decay.
All optimizer hyperparameters that are not explicitly specified are the same as PyTorch defaults.
Each model is trained for up to 200 epochs, with early stopping triggered if validation loss does not improve for 10 consecutive epochs. 
Predictive uncertainty is estimated by sampling weights 10 times for each model forward pass in training and 100 times per forward pass during validation.
To ensure statistical robustness under different weight initializations, each configuration is trained and evaluated across 30 independent fixed random seeds.

\subsection{Metrics}\label{sec:metrics}

The performance of the model is evaluated using the mean squared error (MSE) for point predictions and the continuous ranked probability score (CRPS) \cite{gneiting2007strictly},
\begin{equation*}
    \operatorname{CRPS}(F, x)=\int_{-\infty}^{\infty}(F(y)-1(y-x))^2 d y \, ,
\end{equation*}
for probabilistic forecasts. 
MSE assesses the accuracy of point estimates, while CRPS also takes into account the quality of the predictive uncertainty.
For the real-world datasets, we also compare Mean Absolute Error (MAE) and Mean Absolute Percentage Error (MAPE).
Calibration is tracked through the agreement between predictions and observations in percentiles of the predicted values.
To evaluate the performance of different likelihood distributions, we conduct unpaired t-tests using the Gaussian likelihood distribution as the baseline. 
For each task, we identify the distribution with the highest performance for each criteria. 
We then assess its statistical significance relative to the Gaussian baseline, marking the results with asterisks: $^{*}$ for $p<0.05$, $^{**}$ for $p<0.01$, or $^{***}$ for $p<0.001$. 
In cases where no single distribution significantly outperforms the others (i.e., all candidates show comparable performance without a statistically significant difference), the entry is labeled with an X to indicate the lack of a clear winner.
The explicit means and standard deviations for each experiment are given in Appendix~\ref{app:fullresults}.

\subsection{Initial Test on Univariate Artificial Data}\label{subsec:uniresult}

As a foundational step, identifying an appropriate fixed value for the degrees of freedom ($\nu$) is essential before proceeding with subsequent experiments. Unlike location and scale parameters, the degree of freedom ($\nu$) in a Student-$t$ likelihood directly dictates the distribution's tail behavior and is often difficult to optimize alongside other parameters via moment-matching. Therefore, we treat $\nu$ as a fixed hyperparameter. To determine the optimal value, we conducted a sensitivity analysis across $\nu \in \{2, 5, 7, 10, 15\}$. As shown in Table~\ref{tab:dof_choice}, we observed a significant performance gap between $\nu=2$ and $\nu=5$, while increasing $\nu$ beyond 5 yielded diminishing returns in both CRPS and MSE. Consequently, we selected $\nu=5$ as it provides a robust balance between outlier tolerance and error convergence.

\begin{table}[h]
\centering
\caption{Comparison of different Degrees of Freedom (DOF) settings, averaging 30 seed runs ($n=3,000$, Gaussian Error)}
\label{tab:dof_choice}
\begin{adjustbox}{max width=\linewidth}
\begin{tabular}{lcccc}
\toprule
DOF Setting & CRPS $\downarrow$ & MSE $\downarrow$ & Training Epochs \\
\midrule
2  & 4.54 $\pm$ 0.32 & 424.16 $\pm$ 230.58 & 19.17 $\pm$ 3.59 \\
\textbf{5}  & \textbf{0.14 $\pm$ 0.01} & \textbf{1.02 $\pm$ 0.17} & 105.72 $\pm$ 6.15 \\
7  & 0.14 $\pm$ 0.01 & 1.02 $\pm$ 0.17 & 105.72 $\pm$ 6.15\\
10 & 0.14 $\pm$ 0.01 & 0.99 $\pm$ 0.20 & 107.22 $\pm$ 7.05\\
15 & 0.14 $\pm$ 0.01 & 1.01 $\pm$ 0.18 & 109.17 $\pm$ 6.31\\
\bottomrule
\end{tabular}
\end{adjustbox}
\end{table}

After deciding the degree of freedom, we aim to test the hypothesis that assuming the likelihood distribution matches the noise distribution of the data yields better predictions.
We use the datasets with the noise following Gaussian, Gamma, or Student’s $t$ (with degree of freedom $\nu=5$) distributions to represent normal, skewed, and heavy-tailed noise, with fixed mean $0$ and standard deviation $3$, respectively. 
We further assume the distributional form of the likelihood distribution to belong to one of three different distributions, Gaussian, skew normal, and Student's $t$ distribution (with fixed degree of freedom $\nu=5$), corresponding to normal, skewed, and heavy-tailed noises.
This yields ``matched'' cases for normal, skewed, and heavy-tailed cases, alongside mismatched combinations; we then determine whether the matched cases perform best.
The skewed likelihood distribution is assumed to be skew normal, instead of Gamma, because the one-sided nature of Gamma prevents proper computation of the loss function, which requires the likelihood distribution to be non-vanishing for all real numbers.

\begin{table}[!h]
\caption{Significance table for the initial experiments.
Rows are grouped by noise type (Gaussian, Gamma, Student’s $t$). For each model architecture ($\text{depth}\times \text{width}$) and sample size $n\in\{300, 3{,}000, 30{,}000\}$, the table reports--separately for MSE and CRPS (lower is better)--which assumed predictive distribution attains the best test score. An asterisk indicates the significance of an unpaired t-test comparing the marked choice against each of the other two. Example: ``T***'' under CRPS at $n=3{,}000$ means Student’s $t$ has the lowest mean CRPS and is significantly better than Gaussian and skew-normal at $p<0.001$.\\}\label{tab:table_exp1}
\begin{adjustbox}{max width=\linewidth}
\begin{tabular}{llllllll}
\toprule
    && \multicolumn{2}{c}{$n=300$} & \multicolumn{2}{c}{$n=3{,}000$} & \multicolumn{2}{c}{$n=30{,}000$}\\
    \cmidrule(lr){3-8}
    \rule{0pt}{12pt}
    NOISE& MODEL&&&&&&\\
    TYPE & SHAPE & MSE $\downarrow$ & CRPS $\downarrow$ & MSE $\downarrow$ & CRPS $\downarrow$ & MSE $\downarrow$ & CRPS $\downarrow$ \\
    \midrule
    \\[-6pt]
    Gamma & 1$\times$32 & X & G$^*$ & X & X & T$^{***}$ & T$^{***}$\\
    & 2$\times$16 & T$^*$ & X & T$^{***}$ & T$^{***}$ & T$^{*}$ & X\\
    & 4$\times$8 & T$^*$ & X & T$^{***}$ & T$^{***}$ & T$^{*}$ & X \\
    & 8$\times$4 & T$^*$ & X & X & T$^{***}$ & T$^{***}$ & T$^{***}$ \\
    \midrule
    Gaussian & 1$\times$32 & X & G$^*$ & X & T$^*$ & X & X\\
    & 2$\times$16 & X & G$^*$ & T$^{**}$ & X & T$^{*}$ & X\\
    & 4$\times$8 & G$^*$ & X & T$^{***}$ & T$^{***}$ & X & X \\
    & 8$\times$4 & G$^*$ & G$^*$ & X & T$^{***}$ & T$^{***}$ & T$^{***}$ \\
    \midrule
    Student's $t$ & 1$\times$ 32 & X & X & T$^{*}$ & T$^{*}$ & T$^{***}$ & T$^{***}$\\
    & 2$\times$ 16 & T$^{***}$ & G$^*$ & T$^{***}$ & T$^{***}$ & T$^{***}$ & X\\
    & 4$\times$ 8 & T$^{**}$ & X & T$^{***}$ & T$^{***}$ & X & X \\
    & 8$\times$ 4 & T$^*$ & X & T$^*$ & T$^{***}$ & T$^{***}$ & T$^{***}$ \\
\bottomrule
\\[-6pt]
\multicolumn{8}{l}{
G: Gaussian\ \
S: skew normal\ \
T: Student's $t$\
}\\
\multicolumn{8}{l}{
X: no significant difference between results of the three predictive distributions.
}\\
\multicolumn{8}{l}{
$^{*}$: $p < 0.05$\ \
$^{**}$: $p < 0.01$\ \
$^{***}$: $p < 0.001$.
}
\end{tabular}
\end{adjustbox}
\vspace{-18pt}
\end{table}

We conduct a standard unpaired t-test to determine significance levels of accuracy differences between choices of likelihood distributions.
The condensed results are summarized in \Cref{tab:table_exp1}, while a full results table with the means and standard deviations of the estimated performance metrics can be found in Appendix~\ref{app:univariateresults}. 
For each combination of model shape, noise type, and dataset size, we indicate which likelihood distribution, Gaussian, skew normal, or Student’s $t$, performs best.

When the training set contains only 300 samples and the noise is Gaussian, the Gaussian likelihood distribution achieves the best performance, consistent with the true data-generating process. 
However, as the sample size increases to 3,000 and 30,000, the Student’s $t$ likelihood distribution either significantly outperforms the others (in more than half of the cases) or results in an "X" label, indicating no other distribution could significantly surpass it. 
This holds regardless of the noise type.
Conversely, the skew normal distribution consistently performs poorly, even with asymmetric noise, suggesting that modeling skewness alone is insufficient for improving predictive accuracy. 
We found no significant performance patterns related to model depth or width.

These findings clarify our main question: the likelihood distribution does not need to match the data noise to be optimal. 
As the sample size grows, rare but influential errors become visible and tail behavior dominates the likelihood-based fit; a Student’s $t$ likelihood can absorb such errors, whereas adding skew alone does not show improvement. 
Thus, Student’s $t$ is a robust default choice for the likelihood distribution when the distributional form of the data noise is unknown, while Gaussian remains suitable for small, nearly-Gaussian samples.

\subsection{Results on Multivariate Artificial Data}

\begin{table*}[!h]
\caption{Significance table for the artificial data experiments.
Rows are grouped by noise type (Gaussian, Gamma, Student’s $t$). For each model architecture ($\text{depth}\times \text{width}$), sample size $n\in\{10,000, 50{,}000\}$, and noise scale ($\sigma_{\varepsilon}$) the table reports--separately for MSE and CRPS (lower is better)--which assumed predictive distribution attains the best test score. An asterisk indicates the significance of an unpaired t-test comparing the marked choice against each of the other two. Example: ``T***'' under $n=10,000$ \& $\sigma_{\varepsilon}=3$ and MSE means Student’s $t$ has the lowest mean MSE and is significantly better than Gaussian and skew-normal at $p<0.001$.\\}\label{tab:tab_multi}
\begin{adjustbox}{max width=\textwidth, max height=\textheight, keepaspectratio}
\begin{tabular}{llllllllll}
\hline
    && \multicolumn{2}{c}{$n=10,000$} \& $\sigma_{\varepsilon}=3$ & \multicolumn{2}{c}{$n=10,000$} \& $\sigma_{\varepsilon}=15$ & \multicolumn{2}{c}{$n=50,000$} \& $\sigma_{\varepsilon}=3$ & \multicolumn{2}{c}{$n=50,000$} \& $\sigma_{\varepsilon}=15$\\
    \cline{3-10}
    \rule{0pt}{12pt}
    NOISE& MODEL&&&&&&\\
    TYPE & SHAPE & MSE $\downarrow$ & CRPS $\downarrow$ & MSE $\downarrow$ & CRPS $\downarrow$ & MSE $\downarrow$ & CRPS $\downarrow$ & MSE $\downarrow$ & CRPS $\downarrow$ \\
    \hline
    Gaussian 
    & 2$\times$64 & T$^{***}$ & T$^{***}$ & T$^*$ & T$^*$ & T$^{***}$ & T$^{***}$ & T$^{***}$ & X\\
    & 2$\times$128 & T$^{***}$ & T$^{***}$ & X & G$^{**}$ & T$^{***}$ & T$^{***}$ & X & G$^{***}$\\
    & 4$\times$64 & T$^{***}$ & T$^{***}$ & T$^{***}$ & X & T$^{***}$ & T$^{***}$ & X & G$^{***}$ \\
    & 4$\times$128 & T$^{***}$ & T$^{***}$ & T$^{***}$ & T$^{***}$ & T$^{***}$ & T$^{***}$ & T$^{**}$ & G$^{**}$ \\
    \hline
    Gamma 
    & 2$\times$64 & T$^{***}$ & T$^{***}$ & T$^{***}$ & T$^{***}$ & T$^{***}$ & T$^{***}$ & G$^{***}$ & T$^{***}$ \\
    (skew)
    & 2$\times$128 & T$^{***}$ & T$^{***}$ & T$^{**}$ & T$^{***}$ & T$^{***}$ & T$^{***}$ & G$^{***}$ & T$^{***}$ \\
    & 4$\times$64 & T$^{***}$ & T$^{***}$ & T$^{***}$ & T$^{***}$ & T$^{***}$ & T$^{***}$ & X & T$^{***}$ \\
    & 4$\times$128 & T$^{***}$ & T$^{***}$ & T$^{***}$ & T$^{***}$ & X & T$^{**}$ & X & T$^{***}$ \\
    \hline
    Lognormal
    & 2$\times$64 & G$^{***}$ & T$^{***}$ & G$^{***}$ & T$^{***}$ & X & T$^{**}$ & G$^{***}$ & T$^{***}$ \\
    (skew)
    & 2$\times$128 & G$^{***}$ & T$^{***}$ & G$^{***}$ & T$^{***}$ & G$^{***}$ & T$^{***}$ & X & T$^{***}$ \\ 
    & 4$\times$64 & G$^{***}$ & T$^{***}$ & G$^{***}$ & T$^{***}$ & X & X & X & T$^{***}$ \\
    & 4$\times$128 & G$^{***}$ & X & G$^{***}$ & T$^{***}$ & X & T$^{*}$ & X & X \\
    \hline
    Student's $t$ 
    & 2$\times$64 & X & T$^{***}$ & T$^{*}$ & T$^{***}$ & T$^{*}$ & T$^{**}$ & T$^{***}$ & T$^{***}$ \\
    (heavy-tail)
    & 2$\times$128 & G$^{***}$ & G$^{***}$ & X & T$^{***}$ & X & X & X & T$^{***}$ \\
    & 4$\times$64 & T$^{***}$ & G$^{**}$ & X & T$^{***}$ & T$^{***}$ & T$^{***}$ & T$^{***}$ & T$^{***}$ \\
    & 4$\times$128 & G$^{***}$ & G$^{***}$ & G$^{***}$ & G$^{***}$ & X & X & T$^{***}$ & X \\
    \hline
    Laplace 
    & 2$\times$64 & T$^{***}$ & T$^{***}$ & T$^{**}$ & T$^{***}$ & T$^{***}$ & T$^{***}$ & T$^{*}$ & T$^{***}$ \\
    (heavy-tail)
    & 2$\times$128 & T$^{***}$ & T$^{***}$ & T$^{***}$ & T$^{***}$ & T$^{***}$ & T$^{***}$ & T$^{***}$ & T$^{***}$ \\
    & 4$\times$64 & T$^{***}$ & T$^{***}$ & T$^{***}$ & T$^{***}$ & T$^{**}$ & T$^{***}$ & T$^{**}$ & T$^{***}$ \\
    & 4$\times$128 & X & T$^{*}$ & X & X & T$^{***}$ & T$^{***}$ & T$^{***}$ & T$^{***}$ \\
    \hline
\\[-6pt]
\multicolumn{8}{l}{
G: Gaussian\ \
S: skew normal\ \
T: Student's $t$\
}\\
\multicolumn{8}{l}{
X: no significant difference between results of the three predictive distributions.
}\\
\multicolumn{8}{l}{
$^{*}$: $p < 0.05$\ \
$^{**}$: $p < 0.01$\ \
$^{***}$: $p < 0.001$.
}
\end{tabular}
\end{adjustbox}
\end{table*}

The results from the univariate dataset encourage us to test the new hypothesis that Student's t is a good default choice for the likelihood distribution on a more complex dataset.
The results for the multivariate regression task are summarized in \Cref{tab:tab_multi}, and detailed in Appendix~\ref{app:multivariateresults}. 
In the experiments, the Student's $t$ likelihood distribution consistently achieves superior or competitive results, even when the ground-truth noise belongs to a different distributional family, such as Gaussian, Gamma, or Laplace. 
This also suggests that the heavy-tailed property of the Student's $t$ likelihood distributions provides a robust learning signal that is more critical for model performance than a precise structural match to the aleatoric noise.

However, in the specific case of lognormal noise, characterized by extreme right-skewness, the Gaussian assumption yields slightly lower MSEs, while the Student's $t$-distribution maintains superior CRPSs. 
This discrepancy highlights a fundamental trade-off: while the thin-tailed Gaussian likelihood effectively ignores extreme residuals in the long right tail, the Student’s $t$-distribution attempts to accommodate them. 
To maximize the likelihood of these outliers while maintaining its inherent symmetry, the Student’s $t$ model undergoes a systematic shift in its location parameter. 
This "pulling" effect slightly biases the predictive mean, leading to the MSE disadvantage. 
Nevertheless, the CRPS metric confirms that the Student's $t$ provides a much more accurate quantification of uncertainty by better covering the probability mass in the tail, which the thin-tailed Gaussian effectively ignores. 

When the ground-truth noise itself follows a Student's t distribution, an interesting sample-size effect emerges.
At a lower sample size ($n=10,000$) and low noise scale ($\sigma_{\varepsilon}=3$), the Gaussian assumption occasionally remains competitive or slightly better in MSE. 
This can be attributed to a regularization effect where the simpler Gaussian model avoids over-fitting to sparse tail events in small datasets.
However, as the sample size increases to $n=50,000$ or the noise becomes more prominent, the Student's $t$ likelihood distribution re-establishes its dominance.
This suggests that with sufficient data, the model can more accurately resolve the underlying symmetry of the heavy-tailed noise, allowing it to balance the location parameter without being disproportionately biased by individual points.

Our experiments demonstrate that while specific noise structures like lognormal skewness can introduce localized trade-offs in mean estimation, the Student’s $t$ likelihood distribution consistently emerges as a more robust choice across diverse noise families and sample sizes. 
Its advantage stems from its inherent flexibility to act as a "safe" default, providing stable training signals in the presence of outliers and superior uncertainty quantification. 
These findings justify its use as a reliable default for VI in BNNs, particularly in real-world scenarios where the underlying noise distribution is often unknown and non-Gaussian.

\begin{figure}[!h]
\centering
\includegraphics[height=1.8in]{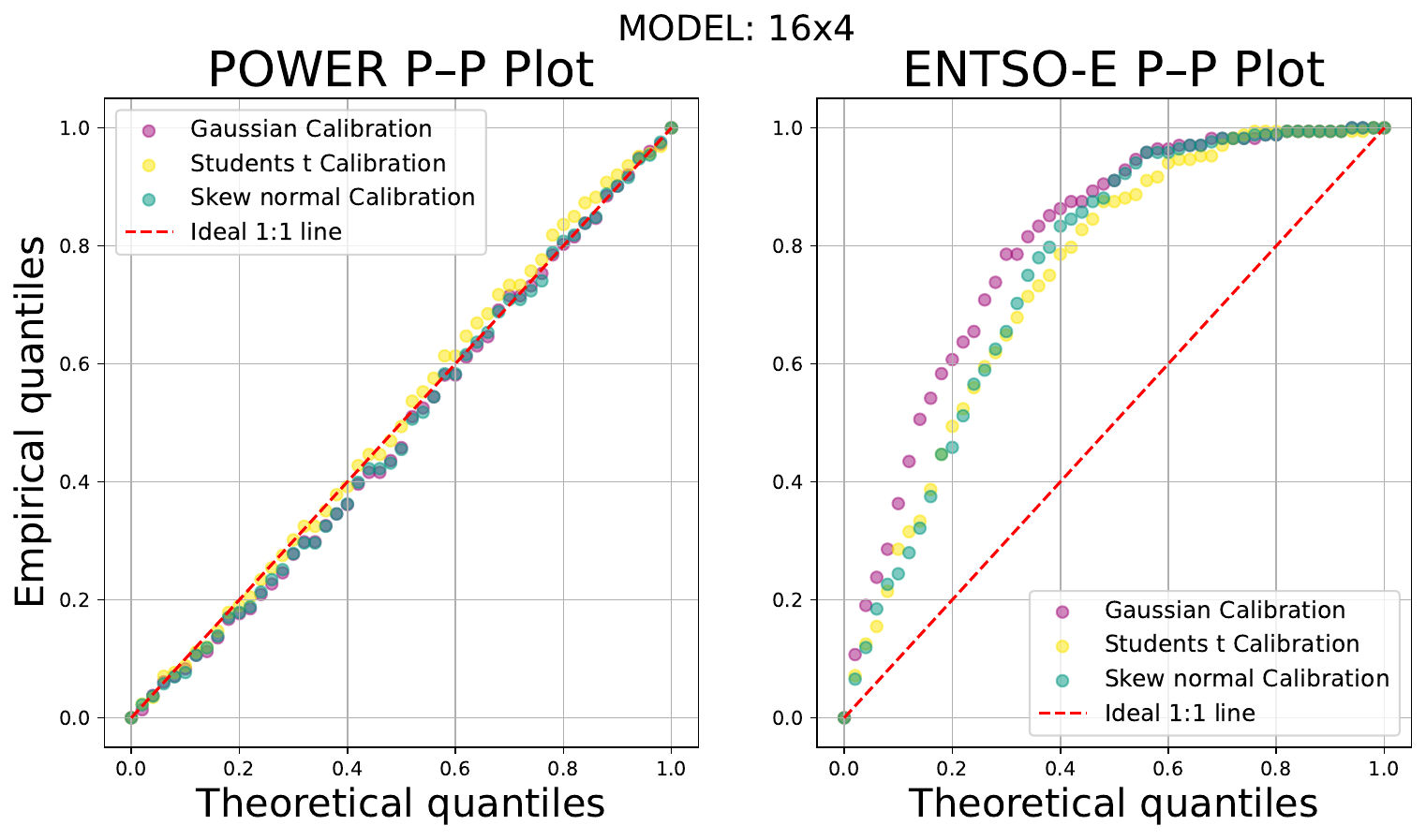}
\caption{Calibration plot the two real-world datasets. This is a plot of model shape: depth 4 and width 16, with all three likelihood distributions.} 
\label{fig:pp-plot}
\end{figure}

\subsection{Results with Real-world Datasets}\label{sec:realworld}

To further solidify our findings that Student's $t$ is a better assumption for the likelihood distribution than the Gaussian in most cases, we performed experiments on the two real-world datasets listed in \Cref{sec:realdata}.
Overall, as can be seen in \Cref{tab:table_real1}, the Student’s $t$ likelihood distribution achieves better performance in most settings, often showing faster convergence as well.
While using the Student’s $t$ likelihood distribution tends to yield longer computation times per epoch, it still requires less time overall in some settings.
The training seems to be more stable, as indicated by the lower standard deviations of the tracked metrics.
In both tasks, the Student’s $t$ likelihood distribution consistently yields lower CRPS and MAE, indicating improved predictive accuracy and uncertainty estimates. 
In some cases, Student’s $t$ distribution produces a lower MAE, while Gaussian achieves a lower MSE. 
This discrepancy may be because MSE penalizes few large errors more heavily, which can disadvantage Student’s $t$ because of its heavy-tailed nature. 
Nevertheless, Student’s $t$ distribution generally seems to offer more robust predictions.
Calibration does not appear to be affected by the likelihood distributions as shown in the P-P plot in \Cref{fig:pp-plot} (plots for different configurations can be found in Appendix~\ref{app:calibrationplots}).
\Cref{fig:entsoe_pred} presents the confidence intervals generated by the three likelihood distributions for a data sample from the ENTSO-E dataset.
It can be observed that the Student’s $t$ likelihood distribution produces narrower intervals, indicating more confident predictions.

\section{Conclusion}\label{sec:conclusion}

Overall, we find that optimal predictive performance does not strictly require the likelihood distribution to match the underlying data-generating noise distribution. 
Instead, the heavy-tailed property of the Student's $t$-distribution provides a more robust learning signal, resulting in more reliable uncertainty quantification, particularly as measured by CRPS.
This could be because the distribution of node values becomes heavy-tailed in later layers, especially with ReLU activations \cite{vladimirova2019understanding,noci2021precise}.
This leads us to the conclusion that Student's $t$-distribution is generally a more robust choice for the likelihood distribution, improving predictive accuracy both point-wise and distributionally, and potentially reducing training time.
Moreover, it is still relatively easy to implement and computationally inexpensive.

In summary, our results imply that it is important to consider more than just the default Gaussian assumption for the likelihood distribution.
Since the ideal choice does not necessarily connected to the distributional form of the data noise (within the distributions that we tested), we find Student's $t$-distribution to be a robust and easy to test alternative.
However, we believe that there is more potential in exploring other distribution types regarding both their predictive performance, but also their corresponding trade-offs in computational efficiency due to increased complexity.

\begin{table}[!h]
\caption{Significance table for POWER ans ENTSO-E dataset. For each model architecture ($\text{depth}\times \text{width}$), the table reports--separately for MAE, MAPE, MSE, CRPS, epoch and training time used (lower is better)---which assumed predictive distribution attains the best test score. An asterisk indicates the significance of an unpaired t-test comparing the marked choice against each of the other two.\\}\label{tab:table_real1}
\begin{adjustbox}{max width=\linewidth}
\begin{tabular}{llllllll}
\toprule
\rule{0pt}{12pt}
    DATASET & MODEL SHAPE & MAE $\downarrow$& MAPE $\downarrow$& MSE $\downarrow$& CRPS $\downarrow$& EPOCH $\downarrow$& TIME $\downarrow$\\
    \midrule
    \\[-6pt]
    POWER &1$\times$ 4 & T$^{***}$ & T$^{***}$ & T$^{***}$ & T$^{***}$ & T$^{***}$ & T$^{*}$ \\
    &1$\times$ 16 & T$^{***}$ & T$^{***}$ & T$^{*}$ & T$^{***}$ & T$^{***}$ & T$^{***}$ \\
    &1$\times$ 32 & T$^{***}$ & T$^{***}$ & T$^{***}$ & T$^{***}$ & T$^{***}$ & T$^{***}$ \\
    \cmidrule(lr){2-8}
    &4$\times$ 4 & T$^{***}$ & T$^{***}$ & T$^{***}$ & T$^{***}$ & X & X \\
    &4$\times$ 16 & T$^{*}$ & $\times$ & X & T$^{***}$ & T$^{***}$ & T$^{***}$ \\
    &4$\times$ 32 & T$^{***}$ & T$^{***}$ & X & T$^{***}$ & T$^{***}$ & T$^{***}$ \\
    \cmidrule(lr){2-8}
    &8$\times$ 4 & T$^{***}$ & T$^{***}$ & G$^{***}$ & T$^{***}$ & X & X \\
    &8$\times$ 16 & T$^{***}$ & T$^{***}$ & T$^{***}$ & T$^{***}$ & X & X \\
    &8$\times$ 32 & T$^{***}$ & T$^{***}$ & T$^{***}$ & T$^{**}$ & T$^{***}$ & T$^{***}$ \\
    \midrule
    \\[-6pt]
    ENTSO-E&1$\times$ 4 & T$^{***}$ & T$^{***}$ & T$^{***}$ &T$^{***}$& $\times$ & X \\
    &1$\times$ 16 & T$^{***}$ & T$^{***}$ & T$^{***}$ &T$^{***}$& $\times$ & X \\
    &1$\times$ 32 & T$^{***}$ & T$^{***}$ & T$^{***}$ &T$^{***}$& T$^{***}$ & X \\
    \cmidrule(lr){2-8}
    &4$\times$ 4 & T$^{***}$ & T$^{***}$ & G$^{***}$ &T$^{***}$& T$^{***}$ & T$^{*}$ \\
    &4$\times$ 16 & T$^{***}$ & T$^{***}$ & X &T$^{*}$& X & X \\
    &4$\times$ 32 & T$^{***}$ & T$^{***}$ & T$^{***}$ &T$^{**}$& T$^{***}$ & T$^{***}$ \\
    \cmidrule(lr){2-8}
    &8$\times$ 4 & T$^{***}$ & T$^{***}$ & G$^{***}$ &X& T$^{**}$ & X \\
    &8$\times$ 16 & T$^{***}$ & T$^{***}$ & T$^{***}$ &T$^{***}$& X & X \\
    &8$\times$ 32 & T$^{***}$ & T$^{***}$ & X &T$^{**}$& T$^{***}$ & X \\
\bottomrule
\\[-6pt]
\multicolumn{8}{l}{
G: Gaussian\ \
S: skew normal\ \
T: Student's $t$\
}\\
\multicolumn{8}{l}{
X: no significant difference between the results of the three predictive distributions.
}\\
\multicolumn{8}{l}{
$^{*}$: $p < 0.05$\ \
$^{**}$: $p < 0.01$\ \
$^{***}$: $p < 0.001$.
}
\end{tabular}
\end{adjustbox}
\vspace{-18pt}
\end{table}

\begin{figure}[!h]
\centering
\includegraphics[height=2.3in]{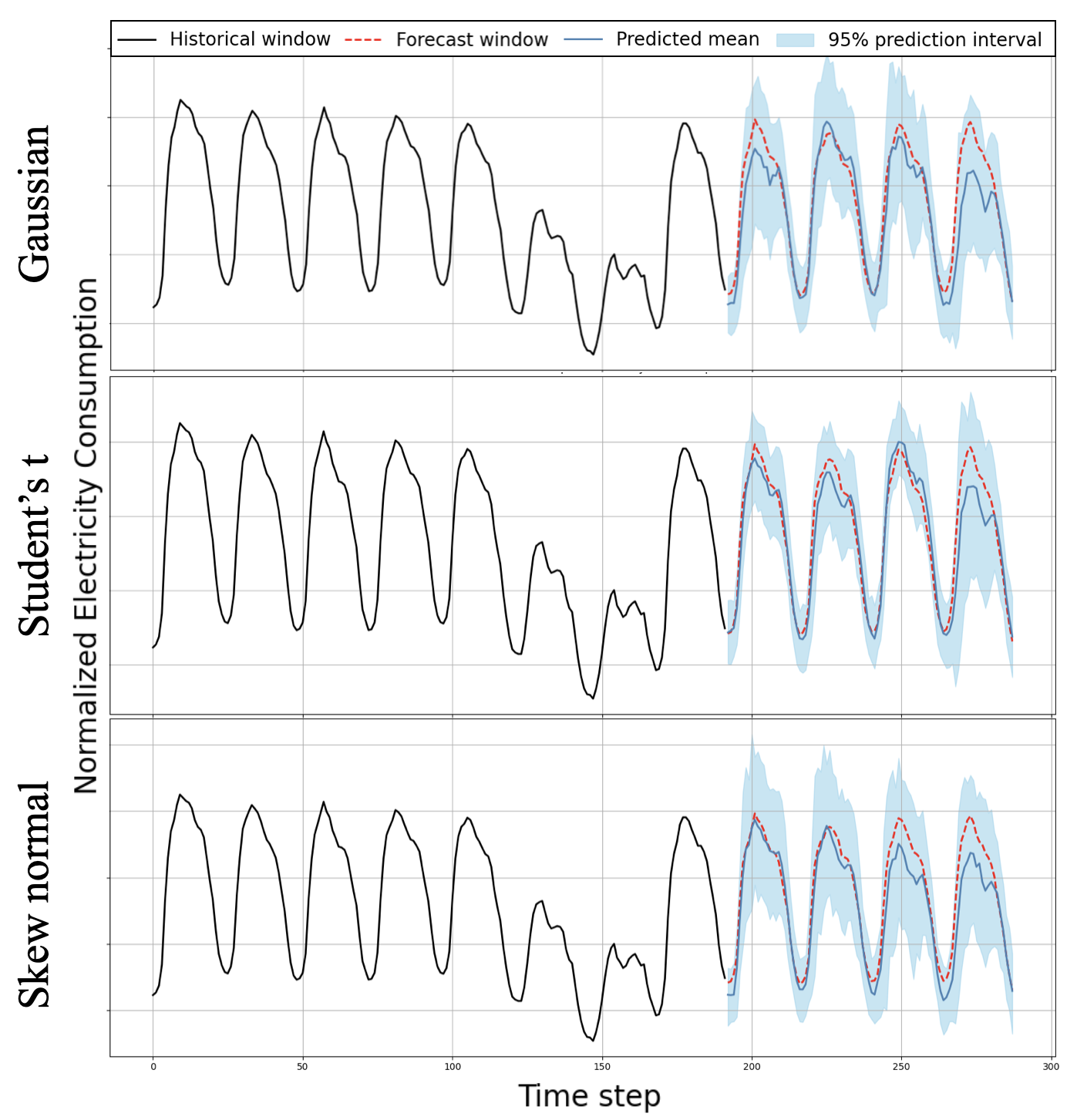}
\caption{Predicted confidence interval of ENTSO-E data set with all three likelihood distributions.} \label{fig:entsoe_pred}
\end{figure}

\newpage

\appendix
\section{Probability density function of distributions}
Here we show the probability density functions of the distributions used in the study.

\subsection{Data generating process} \label{app:noisedists}

Gaussian: $f(x) = \frac{1}{\sqrt{2 \pi \sigma^2}} e^{-\frac{(x-\mu)^2}{2 \sigma^2}}$, where $\mu$ is the mean and $\sigma$ is the standard deviation.

Student's $t$: $f(x) = \frac{\Gamma\left(\frac{\nu+1}{2}\right)}{\sqrt{\pi \nu} \Gamma\left(\frac{\nu}{2}\right)}\left(1+\frac{x^2}{\nu}\right)^{-\frac{\nu+1}{2}}$, where $\nu$ is the degree of freedom.


Gamma: $f(x) = \begin{cases}\frac{1}{\Gamma(\alpha) \theta^\alpha} x^{\alpha-1} e^{-x / \theta} & \text { for } x \geq 0 \\ 0 & \text { otherwise }\end{cases}$, where $\alpha$ is the shape, and $\theta$ is the scale.

Lognormal: $f(x) = \begin{cases}\frac{1}{x \sigma \sqrt{2 \pi}} \exp \left(-\frac{(\ln x-\mu)^2}{2 \sigma^2}\right) & \text { for } x > 0 \\ 0 & \text { otherwise }\end{cases}$, where $\mu$ is the logarithm of location and $\sigma$ is the logarithm of scale.

Laplace: $f(x) = \frac{1}{2 b} \exp \left(-\frac{|x-\mu|}{b}\right)$, where $\mu$ is the location and $b$ is the scale.


\subsection{likelihood distributions} \label{app:preddists}

Gaussian: $f(x)=\frac{1}{\sqrt{2 \pi \sigma^2}} e^{-\frac{(x-\mu)^2}{2 \sigma^2}}$, where the parameters are $\mu$ (mean) and $\sigma$ (standard deviation).

Skew normal: $f(x)=\frac{2}{\omega \sqrt{2 \pi}} e^{-\frac{(x-\xi)^2}{2 \omega^2}} \int_{-\infty}^{\alpha\left(\frac{x-\xi}{\omega}\right)} \frac{1}{\sqrt{2 \pi}} e^{-\frac{t^2}{2}} \mathrm{~d} t$, where the parameters are $\xi$ (location), $\omega$ (scale), and $\alpha$ (shape).

Location-scale $t$ distribution: $f(x)=\frac{\Gamma\left(\frac{\nu+1}{2}\right)}{\Gamma\left(\frac{\nu}{2}\right) \tau \sqrt{\pi \nu}}\left(1+\frac{1}{\nu}\left(\frac{x-\mu}{\tau}\right)^2\right)^{-(\nu+1) / 2}$, where the parameters are $\mu$ (location) and $\tau$ (scale), the degree of freedom $\nu$ is fixed to $5$.

\newpage

\section{Calibration plots for all MLP depths and widths} \label{app:calibrationplots}

\begin{figure}[!h]
\begin{center}
\includegraphics[height=2.0in]{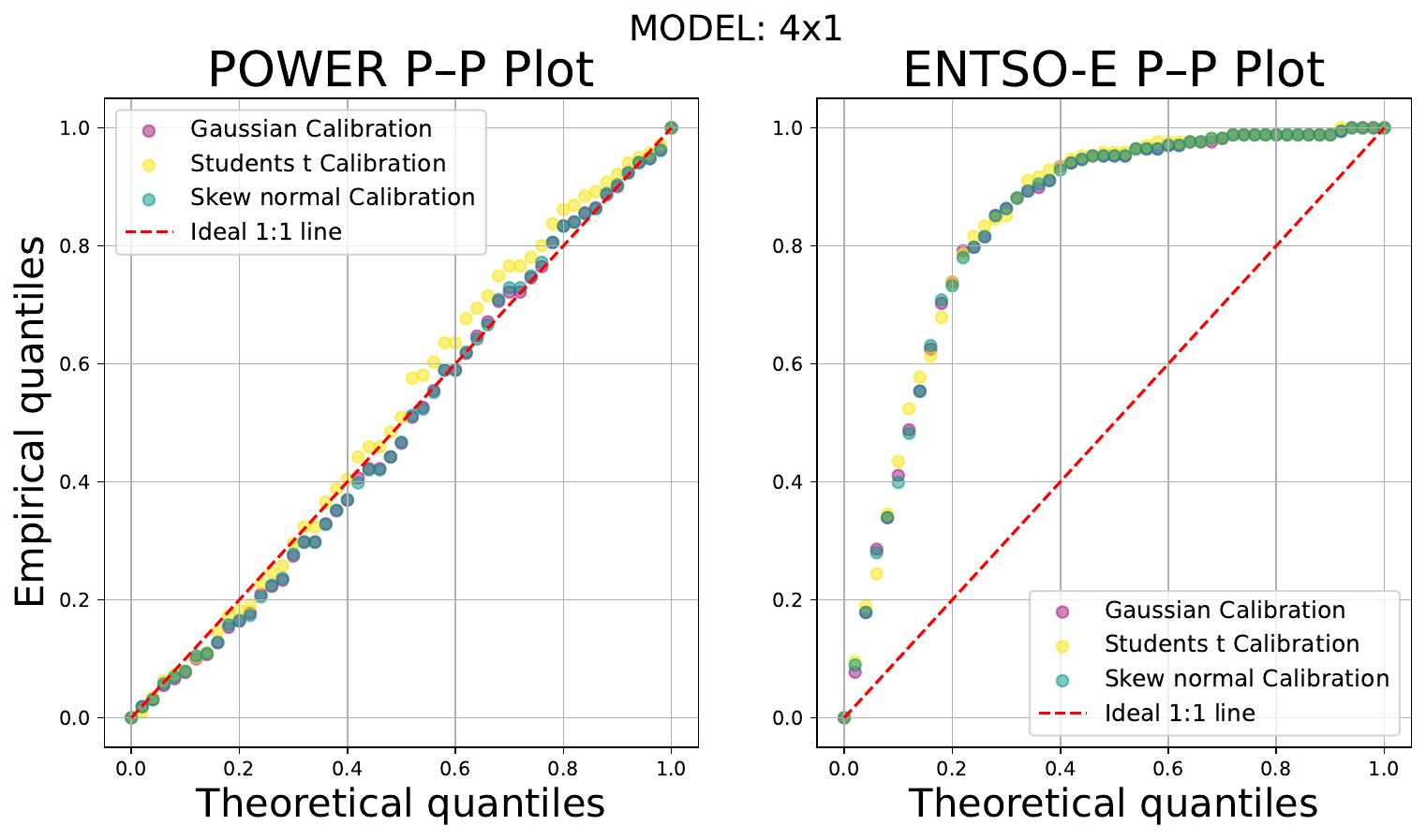}
\caption{Calibration plots for all MLP depth 1 and width 4.} \label{fig:4x1}
\end{center}
\end{figure}

\begin{figure}[!h]
\begin{center}
\includegraphics[height=2.0in]{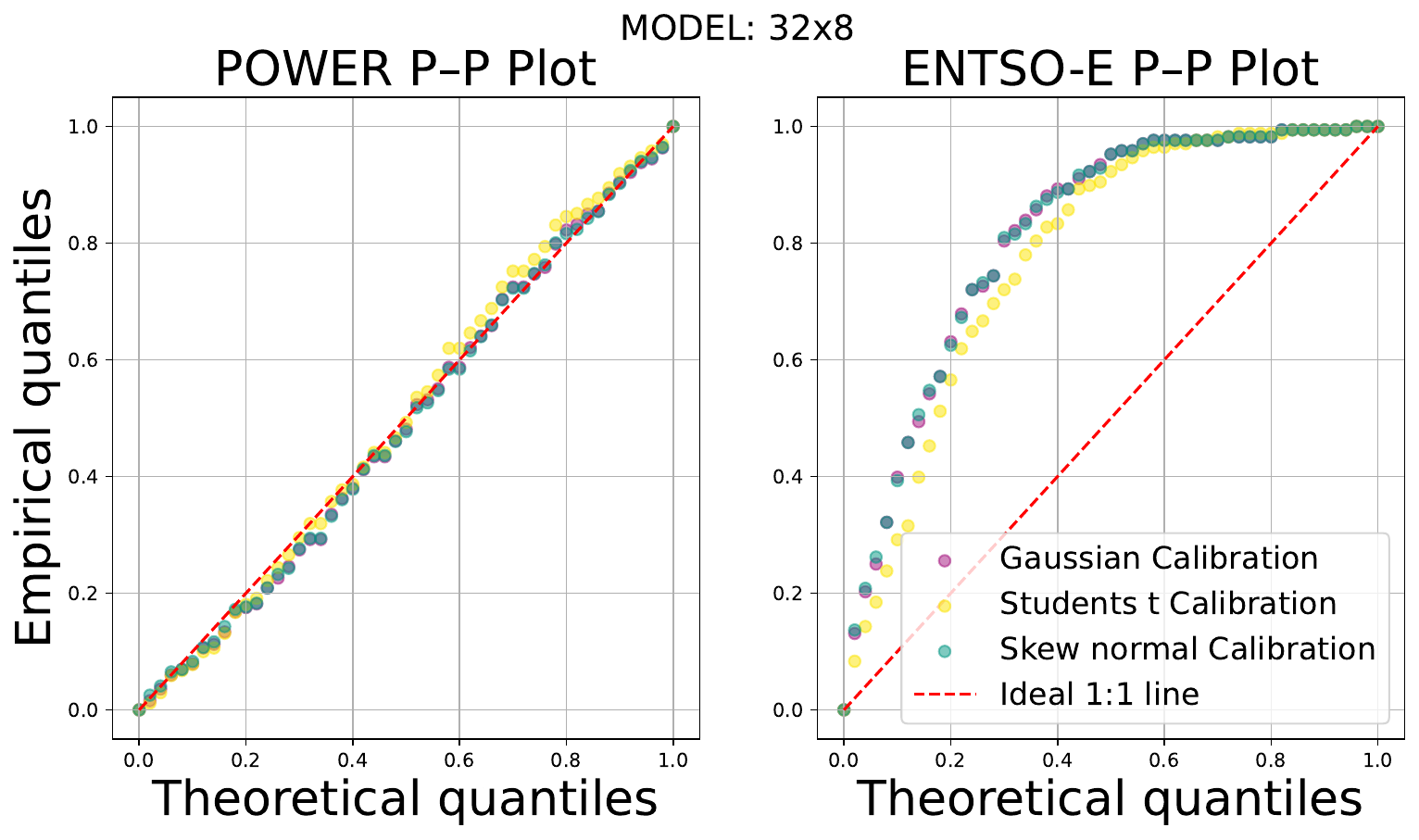}
\caption{Calibration plots for all MLP depth 1 and width 16.} \label{fig:16x1}
\end{center}
\end{figure}

\begin{figure}[!h]
\begin{center}
\includegraphics[height=2.0in]{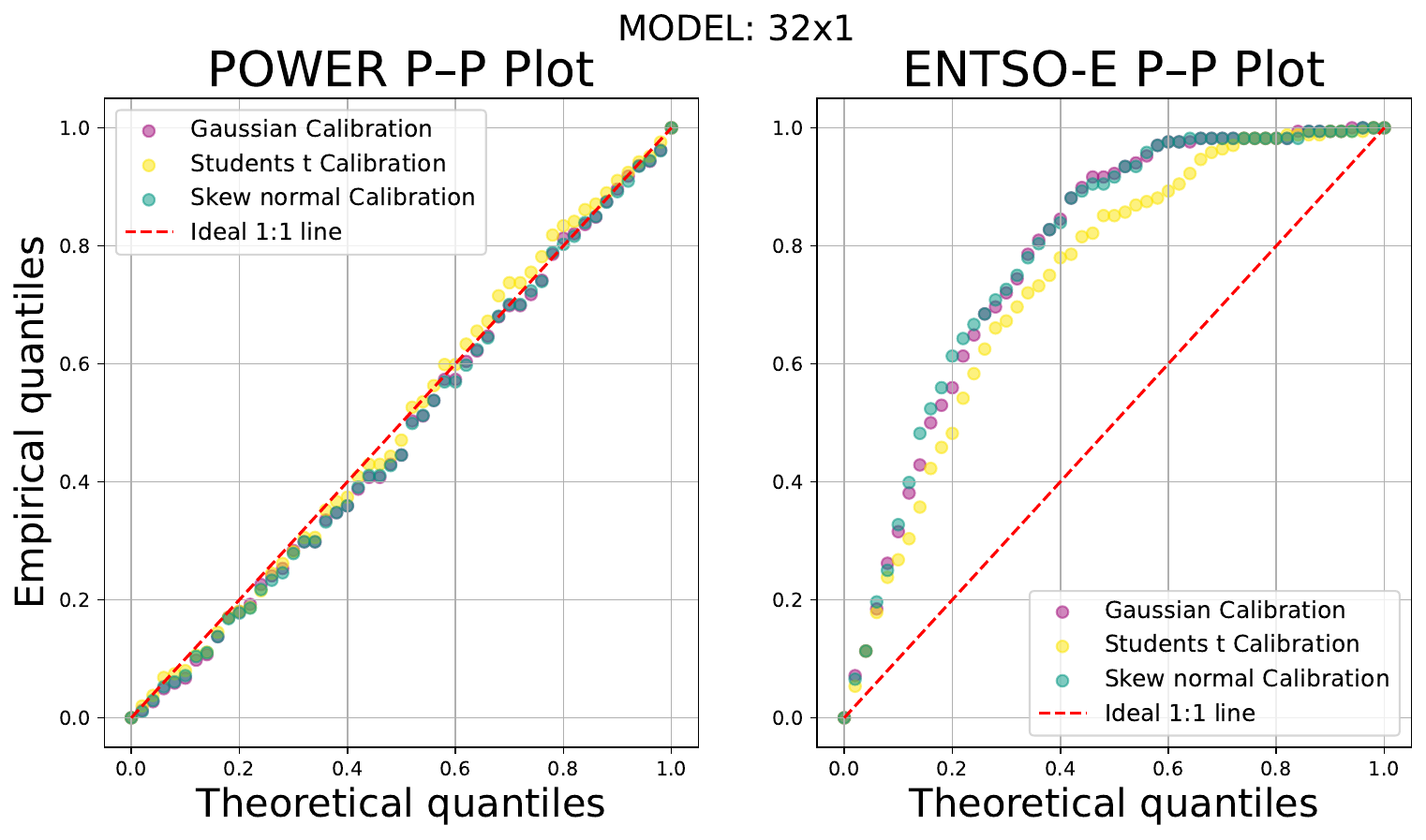}
\caption{Calibration plots for all MLP depth 1 and width 32.} \label{fig:32x1}
\end{center}
\end{figure}

\begin{figure}[!h]
\begin{center}
\includegraphics[height=2.0in]{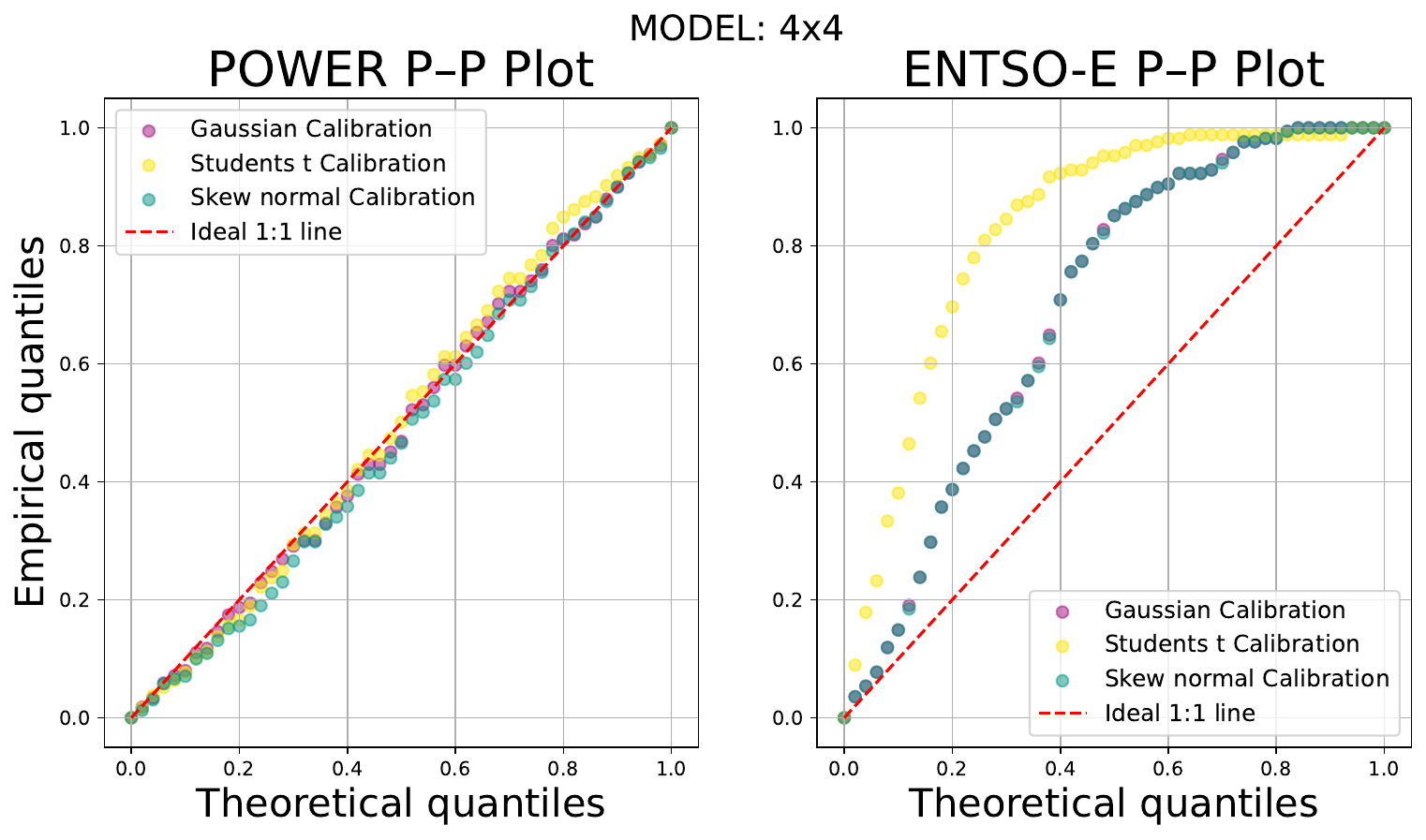}
\caption{Calibration plots for all MLP depth 4 and width 4.} \label{fig:4x4}
\end{center}
\end{figure}

\begin{figure}[!h]
\begin{center}
\includegraphics[height=2.0in]{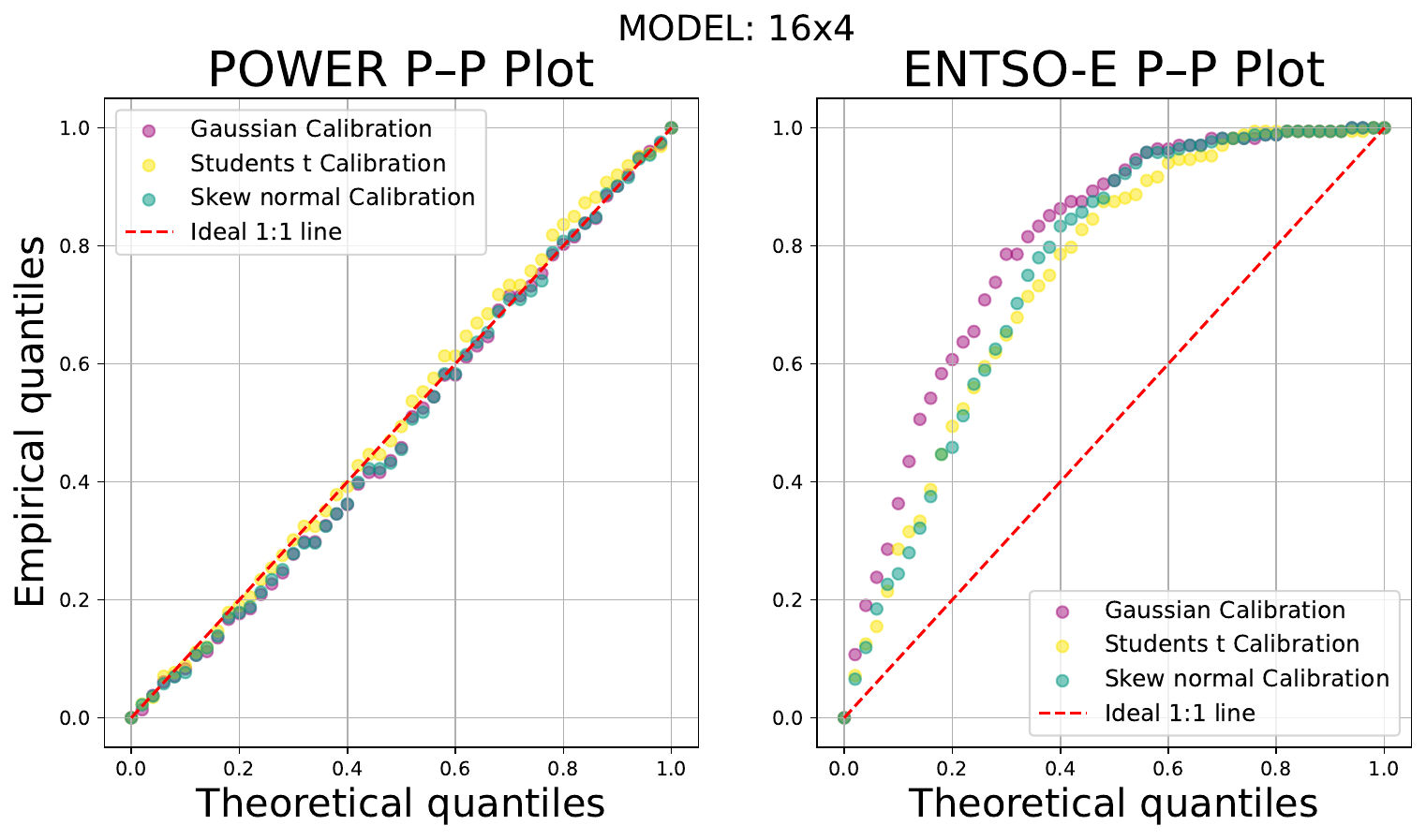}
\caption{Calibration plots for all MLP depth 4 and width 16.} \label{fig:16x4}
\end{center}
\end{figure}

\begin{figure}[!h]
\begin{center}
\includegraphics[height=2.0in]{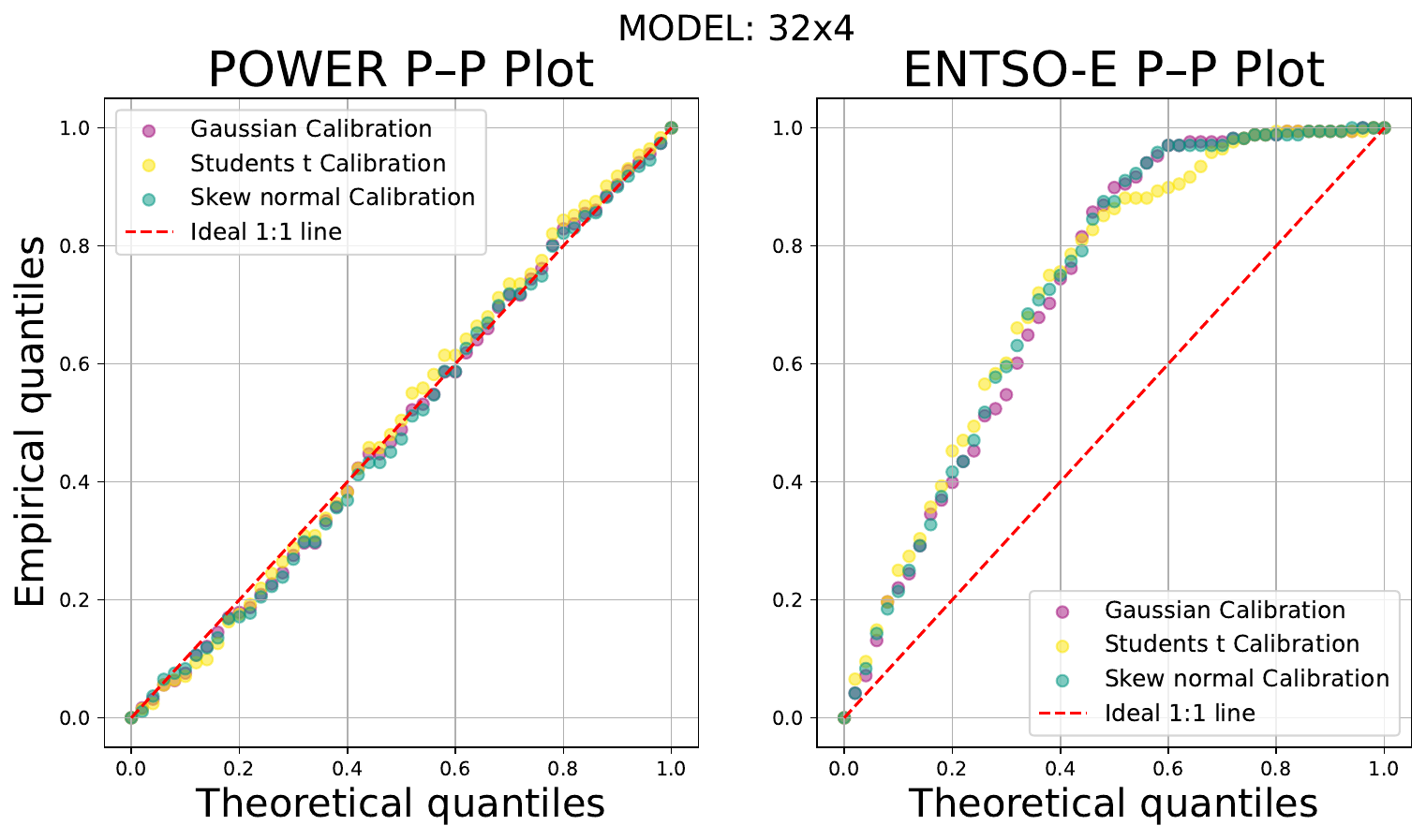}
\caption{Calibration plots for all MLP depth 4 and width 32.} \label{fig:32x4}
\end{center}
\end{figure}

\begin{figure}[!h]
\begin{center}
\includegraphics[height=2.0in]{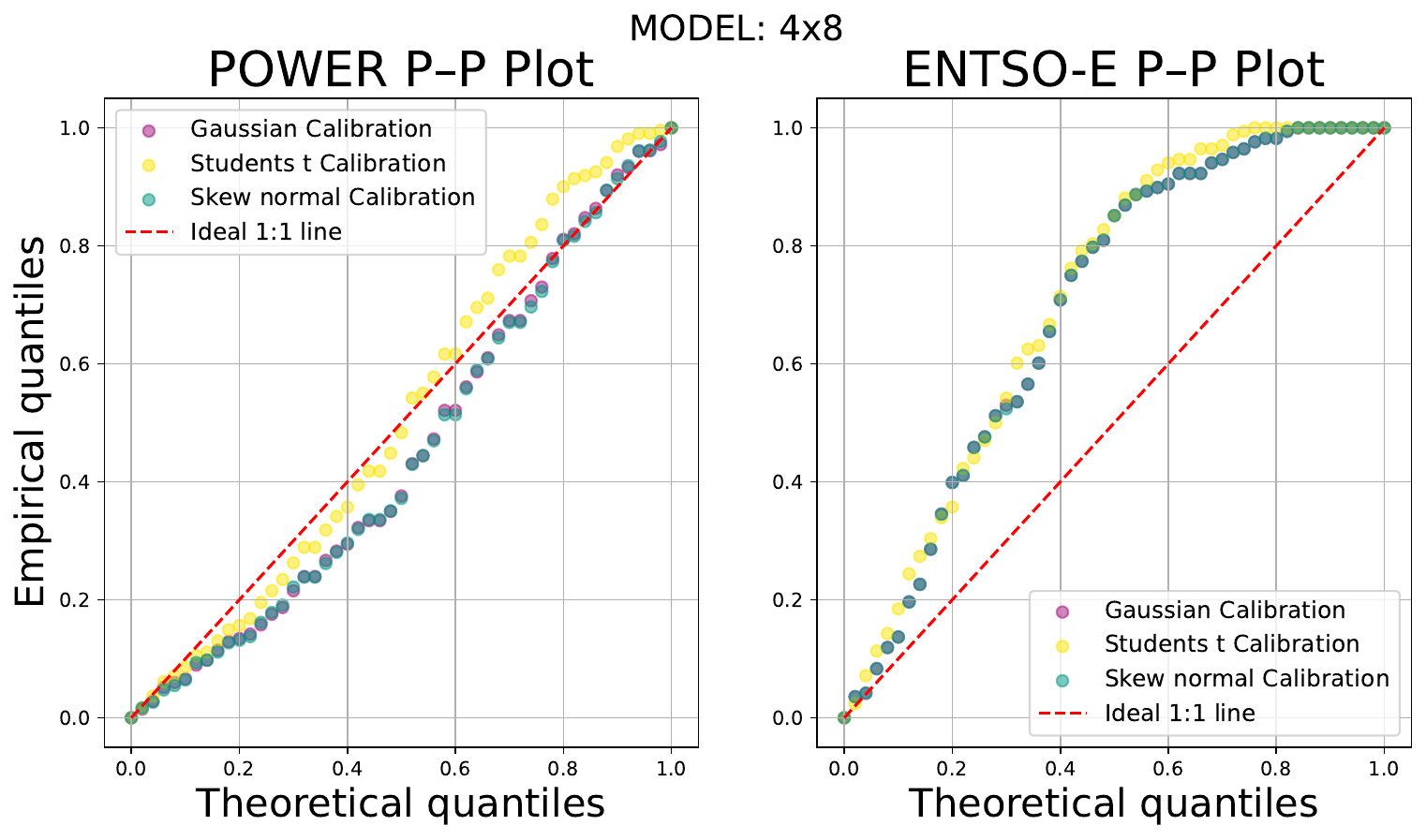}
\caption{Calibration plots for all MLP depth 8 and width 4.} \label{fig:4x8}
\end{center}
\end{figure}

\begin{figure}[!h]
\begin{center}
\includegraphics[height=2.0in]{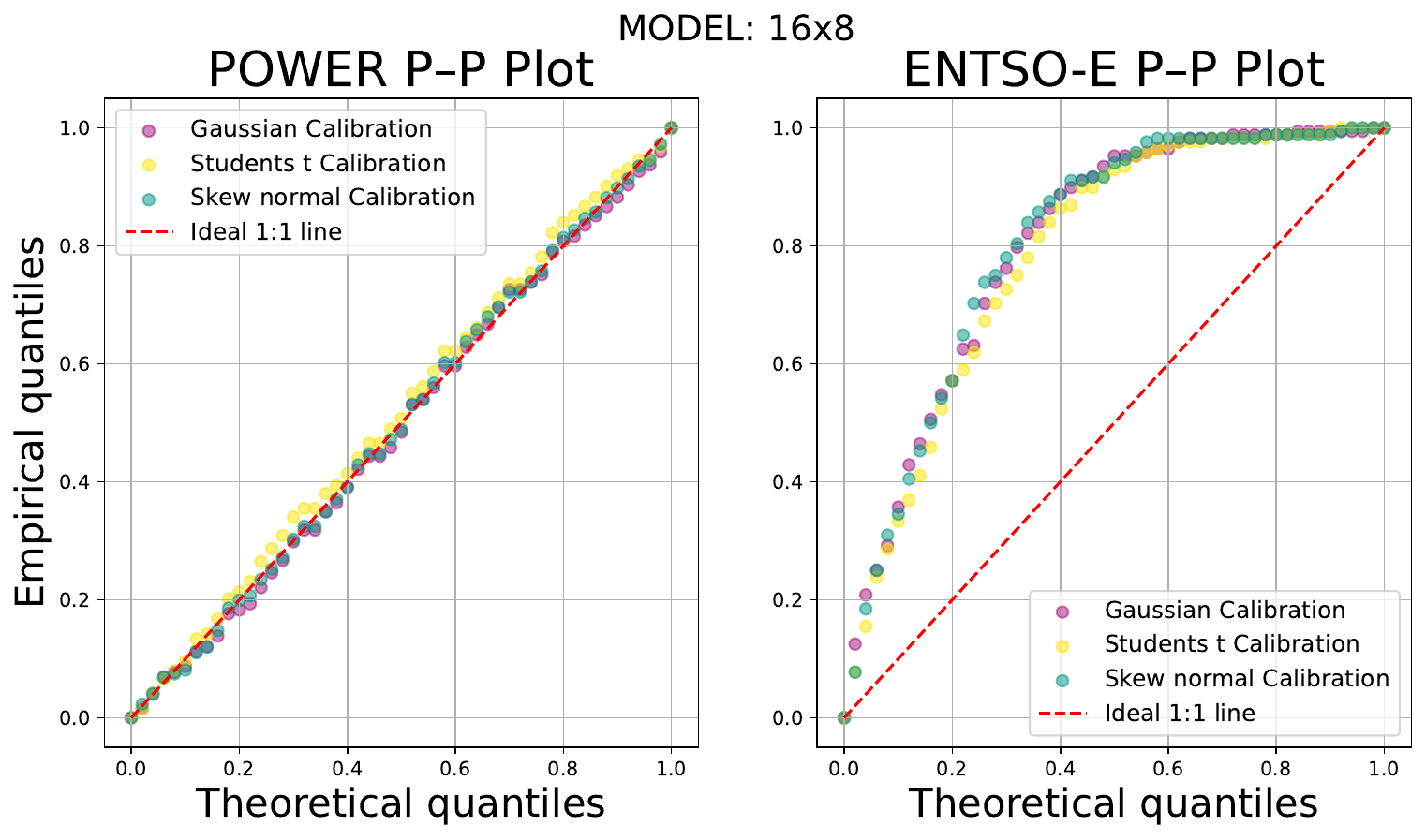}
\caption{Calibration plots for all MLP depth 8 and width 16.} \label{fig:16x8}
\end{center}
\end{figure}

\begin{figure}[!h]
\begin{center}
\includegraphics[height=2.0in]{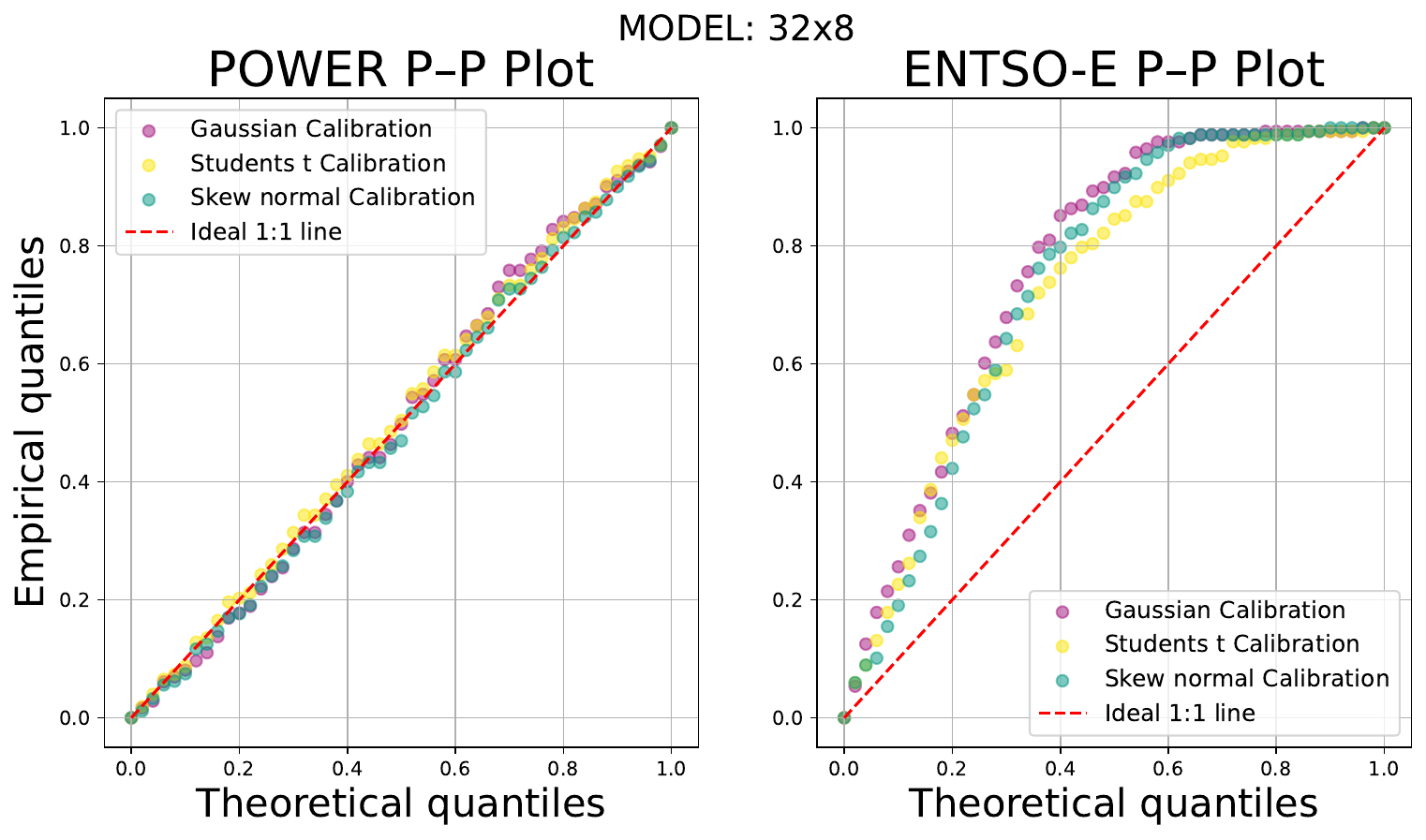}
\caption{Calibration plots for all MLP depth 8 and width 32.} \label{fig:32x8}
\end{center}
\end{figure}

\newpage

\section{Full results of the experiments} \label{app:fullresults}

\subsection{Univariate data results} \label{app:univariateresults}

\begin{table}[H] 
\centering
\begin{adjustbox}{max width=\linewidth, max height=10cm}
\begin{tabular}{lllllll}
\toprule
 &  &  & mse & crps & epoch & time \\
error type & model size & likelihood distribution&  &  &  &  \\
\midrule
\multirow[t]{12}{*}{gamma} & \multirow[t]{3}{*}{1×32} & gaussian & 38.98 ± 10.97 & 2.82 ± 0.26 & 25.47 ± 4.23 & 2.16 ± 0.36 \\
 &  & skew & 44.04 ± 11.86 & 2.98 ± 0.25 & 24.7 ± 4.31 & 2.81 ± 0.5 \\
 &  & student's $t$ & 40.8 ± 11.9 & 3.03 ± 0.27 & 24.1 ± 4.74 & 2.17 ± 0.42 \\
\cline{2-7}
 & \multirow[t]{3}{*}{2×16} & gaussian & 40.85 ± 6.71 & 1.57 ± 0.19 & 33.0 ± 6.61 & 3.82 ± 0.76 \\
 &  & skew & 38.51 ± 5.78 & 1.64 ± 0.22 & 33.23 ± 6.81 & 4.58 ± 0.94 \\
 &  & student's $t$ & 34.86 ± 5.39 & 1.87 ± 0.22 & 29.77 ± 6.09 & 3.47 ± 0.71 \\
\cline{2-7}
 & \multirow[t]{3}{*}{4×8} & gaussian & 55.64 ± 11.19 & 1.77 ± 0.2 & 33.53 ± 6.63 & 5.75 ± 1.14 \\
 &  & skew & 54.12 ± 9.48 & 1.83 ± 0.24 & 33.47 ± 7.32 & 6.44 ± 1.41 \\
 &  & student's $t$ & 49.59 ± 6.23 & 1.86 ± 0.18 & 31.83 ± 6.19 & 5.43 ± 1.05 \\
\cline{2-7}
 & \multirow[t]{3}{*}{8×4} & gaussian & 71.84 ± 31.5 & 2.11 ± 0.39 & 36.53 ± 7.11 & 10.45 ± 2.04 \\
 &  & skew & 76.05 ± 41.28 & 2.21 ± 0.47 & 38.37 ± 8.7 & 11.57 ± 2.61 \\
 &  & student's $t$ & 55.09 ± 9.52 & 2.11 ± 0.33 & 38.3 ± 6.08 & 11.07 ± 1.77 \\
\cline{1-7} \cline{2-7}
\multirow[t]{12}{*}{gaussian} & \multirow[t]{3}{*}{1×32} & gaussian & 97.79 ± 16.76 & 2.87 ± 0.24 & 26.07 ± 5.52 & 2.22 ± 0.45 \\
 &  & skew & 106.15 ± 18.25 & 3.04 ± 0.22 & 24.93 ± 5.35 & 2.89 ± 0.6 \\
 &  & student's $t$ & 94.12 ± 16.03 & 2.99 ± 0.22 & 25.73 ± 5.43 & 2.34 ± 0.48 \\
\cline{2-7}
 & \multirow[t]{3}{*}{2×16} & gaussian & 126.58 ± 13.84 & 2.01 ± 0.11 & 27.97 ± 4.68 & 3.25 ± 0.54 \\
 &  & skew & 129.79 ± 12.91 & 2.07 ± 0.13 & 28.47 ± 4.92 & 4.04 ± 0.68 \\
 &  & student's $t$ & 122.24 ± 15.38 & 2.16 ± 0.14 & 27.1 ± 5.72 & 3.14 ± 0.64 \\
\cline{2-7}
 & \multirow[t]{3}{*}{4×8} & gaussian & 162.41 ± 13.4 & 2.3 ± 0.13 & 28.93 ± 7.41 & 4.81 ± 1.23 \\
 &  & skew & 168.45 ± 13.64 & 2.36 ± 0.14 & 29.3 ± 8.12 & 5.7 ± 1.58 \\
 &  & student's $t$ & 170.54 ± 14.85 & 2.4 ± 0.15 & 27.73 ± 7.32 & 4.72 ± 1.24 \\
\cline{2-7}
 & \multirow[t]{3}{*}{8×4} & gaussian & 180.87 ± 14.33 & 2.48 ± 0.21 & 35.07 ± 6.76 & 10.06 ± 1.95 \\
 &  & skew & 191.24 ± 22.82 & 2.58 ± 0.23 & 31.43 ± 7.02 & 9.53 ± 2.15 \\
 &  & student's $t$ & 187.52 ± 13.22 & 2.59 ± 0.21 & 31.93 ± 5.89 & 8.93 ± 1.65 \\
\cline{1-7} \cline{2-7}
\multirow[t]{12}{*}{student's $t$} & \multirow[t]{3}{*}{1×32} & gaussian & 101.88 ± 16.63 & 2.96 ± 0.23 & 25.83 ± 4.34 & 2.29 ± 0.36 \\
 &  & skew & 94.54 ± 16.14 & 3.05 ± 0.24 & 26.77 ± 5.0 & 2.99 ± 0.56 \\
 &  & student's $t$ & 88.41 ± 15.52 & 3.1 ± 0.24 & 25.5 ± 4.93 & 2.28 ± 0.42 \\
\cline{2-7}
 & \multirow[t]{3}{*}{2×16} & gaussian & 142.03 ± 15.17 & 2.03 ± 0.15 & 28.87 ± 5.58 & 3.27 ± 0.63 \\
 &  & skew & 136.47 ± 15.23 & 2.1 ± 0.16 & 29.2 ± 5.39 & 4.18 ± 0.77 \\
 &  & student's $t$ & 117.25 ± 15.59 & 2.17 ± 0.23 & 27.47 ± 5.93 & 3.27 ± 0.7 \\
\cline{2-7}
 & \multirow[t]{3}{*}{4×8} & gaussian & 176.5 ± 16.82 & 2.29 ± 0.19 & 31.8 ± 5.53 & 5.39 ± 0.92 \\
 &  & skew & 169.15 ± 15.65 & 2.3 ± 0.2 & 33.5 ± 6.76 & 6.7 ± 1.33 \\
 &  & student's $t$ & 159.63 ± 9.43 & 2.36 ± 0.21 & 29.2 ± 4.66 & 4.93 ± 0.77 \\
\cline{2-7}
 & \multirow[t]{3}{*}{8×4} & gaussian & 183.29 ± 34.15 & 2.5 ± 0.32 & 34.9 ± 7.83 & 9.69 ± 2.15 \\
 &  & skew & 207.54 ± 68.11 & 2.73 ± 0.45 & 34.27 ± 7.92 & 10.81 ± 2.48 \\
 &  & student's $t$ & 167.03 ± 20.22 & 2.59 ± 0.33 & 34.1 ± 5.19 & 9.89 ± 1.48 \\
\bottomrule
\end{tabular}
\end{adjustbox}
\caption{Results for artificial univariate data with error scale $3$, $n=300$}
\label{tab:placeholder}
\end{table}

\begin{table}[H]
\centering
\begin{adjustbox}{max width=\linewidth, max height=12cm}
    \begin{tabular}{lllllll}
\toprule
 &  &  & mse & crps & epoch & time \\
error type & model size & likelihood distribution &  &  &  &  \\
\midrule
\multirow[t]{12}{*}{gamma} & \multirow[t]{3}{*}{1×32} & gaussian & 0.89 ± 0.2 & 0.13 ± 0.01 & 106.8 ± 9.28 & 90.83 ± 7.85 \\
 &  & skew & 2.64 ± 3.12 & 0.17 ± 0.07 & 98.7 ± 20.96 & 112.42 ± 23.78 \\
 &  & student's $t$ & 1.0 ± 1.1 & 0.13 ± 0.03 & 102.83 ± 10.87 & 93.43 ± 10.15 \\
\cline{2-7}
 & \multirow[t]{3}{*}{2×16} & gaussian & 0.68 ± 0.08 & 0.12 ± 0.01 & 137.3 ± 12.34 & 153.94 ± 13.89 \\
 &  & skew & 1.97 ± 2.23 & 0.16 ± 0.07 & 147.57 ± 21.46 & 207.63 ± 30.15 \\
 &  & student's $t$ & 0.61 ± 0.03 & 0.11 ± 0.0 & 117.8 ± 8.96 & 139.32 ± 10.69 \\
\cline{2-7}
 & \multirow[t]{3}{*}{4×8} & gaussian & 121.36 ± 5.02 & 1.55 ± 0.02 & 28.17 ± 5.7 & 47.82 ± 9.72 \\
 &  & skew & 124.29 ± 2.12 & 1.57 ± 0.01 & 27.07 ± 4.37 & 54.59 ± 8.82 \\
 &  & student's $t$ & 17.62 ± 35.02 & 0.37 ± 0.44 & 134.37 ± 49.62 & 239.05 ± 88.13 \\
\cline{2-7}
 & \multirow[t]{3}{*}{8×4} & gaussian & 129.9 ± 0.78 & 1.59 ± 0.0 & 25.8 ± 2.87 & 74.28 ± 8.26 \\
 &  & skew & 129.94 ± 0.77 & 1.6 ± 0.0 & 26.5 ± 2.99 & 80.8 ± 9.06 \\
 &  & student's $t$ & 129.7 ± 0.63 & 1.58 ± 0.0 & 28.53 ± 4.06 & 82.27 ± 11.58 \\
\cline{1-7} \cline{2-7}
\multirow[t]{12}{*}{gaussian} & \multirow[t]{3}{*}{1×32} & gaussian & 1.76 ± 2.73 & 0.16 ± 0.05 & 110.33 ± 16.08 & 93.05 ± 13.54 \\
 &  & skew & 6.5 ± 6.58 & 0.25 ± 0.12 & 97.4 ± 24.89 & 109.41 ± 27.84 \\
 &  & student's $t$ & 1.02 ± 0.19 & 0.14 ± 0.01 & 111.37 ± 13.15 & 100.95 ± 12.02 \\
\cline{2-7}
 & \multirow[t]{3}{*}{2×16} & gaussian & 0.89 ± 0.18 & 0.13 ± 0.01 & 140.13 ± 12.17 & 158.9 ± 13.8 \\
 &  & skew & 7.77 ± 7.06 & 0.29 ± 0.14 & 125.3 ± 31.88 & 179.62 ± 45.83 \\
 &  & student's $t$ & 0.81 ± 0.05 & 0.13 ± 0.0 & 124.93 ± 9.54 & 151.24 ± 11.5 \\
\cline{2-7}
 & \multirow[t]{3}{*}{4×8} & gaussian & 133.37 ± 37.6 & 1.6 ± 0.29 & 37.83 ± 22.79 & 63.64 ± 38.32 \\
 &  & skew & 154.88 ± 3.88 & 1.74 ± 0.02 & 28.17 ± 5.22 & 56.51 ± 10.44 \\
 &  & student's $t$ & 11.08 ± 16.66 & 0.32 ± 0.31 & 143.87 ± 42.12 & 256.09 ± 74.98 \\
\cline{2-7}
 & \multirow[t]{3}{*}{8×4} & gaussian & 163.34 ± 1.16 & 1.78 ± 0.01 & 27.37 ± 3.16 & 79.21 ± 9.18 \\
 &  & skew & 163.57 ± 1.16 & 1.79 ± 0.01 & 27.5 ± 3.19 & 84.48 ± 9.85 \\
 &  & student's $t$ & 163.45 ± 1.37 & 1.77 ± 0.01 & 28.37 ± 3.36 & 82.8 ± 9.8 \\
\cline{1-7} \cline{2-7}
\multirow[t]{12}{*}{student's $t$} & \multirow[t]{3}{*}{1×32} & gaussian & 2.59 ± 3.79 & 0.17 ± 0.08 & 104.37 ± 23.13 & 87.2 ± 19.32 \\
 &  & skew & 2.81 ± 3.64 & 0.18 ± 0.07 & 103.47 ± 16.19 & 116.86 ± 18.36 \\
 &  & student's $t$ & 1.06 ± 0.27 & 0.14 ± 0.02 & 108.93 ± 10.52 & 97.79 ± 9.29 \\
\cline{2-7}
 & \multirow[t]{3}{*}{2×16} & gaussian & 0.78 ± 0.04 & 0.13 ± 0.0 & 143.83 ± 8.02 & 169.12 ± 9.41 \\
 &  & skew & 2.24 ± 3.11 & 0.17 ± 0.07 & 153.77 ± 24.88 & 213.34 ± 34.98 \\
 &  & student's $t$ & 0.74 ± 0.04 & 0.12 ± 0.0 & 127.73 ± 7.83 & 147.12 ± 9.23 \\
\cline{2-7}
 & \multirow[t]{3}{*}{4×8} & gaussian & 145.52 ± 14.85 & 1.68 ± 0.08 & 28.8 ± 9.18 & 48.55 ± 15.44 \\
 &  & skew & 151.54 ± 3.04 & 1.71 ± 0.02 & 27.73 ± 5.25 & 53.5 ± 10.14 \\
 &  & student's $t$ & 18.83 ± 30.09 & 0.42 ± 0.41 & 126.87 ± 43.13 & 223.35 ± 75.86 \\
\cline{2-7}
 & \multirow[t]{3}{*}{8×4} & gaussian & 157.75 ± 0.96 & 1.74 ± 0.01 & 28.77 ± 4.22 & 82.33 ± 11.96 \\
 &  & skew & 157.77 ± 1.0 & 1.74 ± 0.01 & 29.73 ± 4.31 & 89.82 ± 12.88 \\
 &  & student's $t$ & 157.33 ± 0.88 & 1.73 ± 0.01 & 29.6 ± 3.65 & 84.7 ± 10.32 \\
\bottomrule
\end{tabular}
\end{adjustbox}
\caption{Results for artificial univariate data with error scale $3$, $n=3,000$}
\label{tab:placeholder}
\end{table}

\begin{table}[H]
\centering
\begin{adjustbox}{max width=\linewidth, max height=12cm}
\begin{tabular}{lllllll}
\toprule
 &  &  & mse & crps & epoch & time \\
error type & model size & likelihood distribution &  &  &  &  \\
\midrule
\multirow[t]{12}{*}{gamma} & \multirow[t]{3}{*}{1×32} & gaussian & 0.72 ± 0.03 & 0.12 ± 0.0 & 44.4 ± 6.26 & 386.23 ± 53.69 \\
 &  & skew & 0.72 ± 0.03 & 0.11 ± 0.0 & 47.47 ± 5.97 & 522.77 ± 66.2 \\
 &  & student's $t$ & 0.69 ± 0.03 & 0.11 ± 0.0 & 42.1 ± 7.3 & 378.66 ± 67.53 \\
\cline{2-7}
 & \multirow[t]{3}{*}{2×16} & gaussian & 0.69 ± 0.03 & 0.11 ± 0.0 & 42.4 ± 5.31 & 488.09 ± 60.8 \\
 &  & skew & 0.7 ± 0.03 & 0.11 ± 0.0 & 43.5 ± 5.11 & 615.32 ± 71.86 \\
 &  & student's $t$ & 0.67 ± 0.02 & 0.11 ± 0.0 & 39.9 ± 4.88 & 459.9 ± 55.92 \\
\cline{2-7}
 & \multirow[t]{3}{*}{4×8} & gaussian & 0.69 ± 0.04 & 0.11 ± 0.0 & 49.9 ± 4.31 & 842.47 ± 71.78 \\
 &  & skew & 0.67 ± 0.02 & 0.11 ± 0.0 & 55.77 ± 4.5 & 1097.12 ± 88.23 \\
 &  & student's $t$ & 0.67 ± 0.02 & 0.11 ± 0.0 & 43.33 ± 4.07 & 757.86 ± 69.67 \\
\cline{2-7}
 & \multirow[t]{3}{*}{8×4} & gaussian & 142.21 ± 0.29 & 1.66 ± 0.0 & 18.43 ± 2.74 & 526.49 ± 78.38 \\
 &  & skew & 142.21 ± 0.34 & 1.66 ± 0.0 & 18.17 ± 2.79 & 551.58 ± 83.65 \\
 &  & student's $t$ & 141.69 ± 0.35 & 1.64 ± 0.0 & 19.37 ± 3.51 & 542.0 ± 98.14 \\
\cline{1-7} \cline{2-7}
\multirow[t]{12}{*}{gaussian} & \multirow[t]{3}{*}{1×32} & gaussian & 0.68 ± 0.02 & 0.12 ± 0.0 & 44.6 ± 5.68 & 377.66 ± 49.34 \\
 &  & skew & 0.68 ± 0.02 & 0.12 ± 0.0 & 49.07 ± 7.51 & 556.25 ± 84.91 \\
 &  & student's $t$ & 0.68 ± 0.02 & 0.12 ± 0.0 & 41.47 ± 5.08 & 373.59 ± 45.48 \\
\cline{2-7}
 & \multirow[t]{3}{*}{2×16} & gaussian & 0.67 ± 0.02 & 0.12 ± 0.0 & 44.33 ± 5.79 & 495.76 ± 65.36 \\
 &  & skew & 0.68 ± 0.03 & 0.12 ± 0.0 & 46.5 ± 5.53 & 657.71 ± 78.11 \\
 &  & student's $t$ & 0.66 ± 0.02 & 0.12 ± 0.0 & 40.77 ± 4.41 & 478.03 ± 51.7 \\
\cline{2-7}
 & \multirow[t]{3}{*}{4×8} & gaussian & 0.66 ± 0.03 & 0.12 ± 0.0 & 48.73 ± 6.55 & 833.49 ± 111.83 \\
 &  & skew & 0.67 ± 0.03 & 0.12 ± 0.0 & 58.2 ± 6.7 & 1121.43 ± 129.78 \\
 &  & student's $t$ & 0.66 ± 0.02 & 0.12 ± 0.0 & 45.7 ± 6.33 & 793.38 ± 109.16 \\
\cline{2-7}
 & \multirow[t]{3}{*}{8×4} & gaussian & 140.88 ± 0.27 & 1.65 ± 0.0 & 19.23 ± 2.53 & 532.4 ± 68.66 \\
 &  & skew & 140.9 ± 0.31 & 1.65 ± 0.0 & 19.63 ± 2.66 & 600.23 ± 79.58 \\
 &  & student's $t$ & 140.55 ± 0.25 & 1.63 ± 0.0 & 18.97 ± 3.05 & 551.4 ± 87.89 \\
\cline{1-7} \cline{2-7}
\multirow[t]{12}{*}{student's $t$} & \multirow[t]{3}{*}{1×32} & gaussian & 0.69 ± 0.02 & 0.12 ± 0.0 & 45.77 ± 7.43 & 398.67 ± 64.43 \\
 &  & skew & 0.7 ± 0.02 & 0.12 ± 0.0 & 46.4 ± 6.37 & 508.85 ± 70.11 \\
 &  & student's $t$ & 0.67 ± 0.03 & 0.11 ± 0.0 & 41.1 ± 5.92 & 358.3 ± 50.34 \\
\cline{2-7}
 & \multirow[t]{3}{*}{2×16} & gaussian & 0.66 ± 0.02 & 0.11 ± 0.0 & 40.27 ± 4.18 & 452.12 ± 47.01 \\
 &  & skew & 0.67 ± 0.03 & 0.11 ± 0.0 & 48.67 ± 6.36 & 674.74 ± 86.59 \\
 &  & student's $t$ & 0.64 ± 0.02 & 0.11 ± 0.0 & 38.9 ± 4.85 & 455.37 ± 57.48 \\
\cline{2-7}
 & \multirow[t]{3}{*}{4×8} & gaussian & 0.64 ± 0.02 & 0.11 ± 0.0 & 48.83 ± 4.07 & 836.47 ± 68.83 \\
 &  & skew & 0.69 ± 0.04 & 0.12 ± 0.0 & 52.43 ± 3.42 & 1029.91 ± 69.36 \\
 &  & student's $t$ & 0.65 ± 0.03 & 0.11 ± 0.0 & 42.9 ± 4.76 & 752.51 ± 82.26 \\
\cline{2-7}
 & \multirow[t]{3}{*}{8×4} & gaussian & 142.73 ± 0.27 & 1.66 ± 0.0 & 20.13 ± 3.45 & 556.43 ± 95.02 \\
 &  & skew & 142.72 ± 0.27 & 1.66 ± 0.0 & 19.47 ± 3.24 & 597.61 ± 100.06 \\
 &  & student's $t$ & 142.23 ± 0.26 & 1.64 ± 0.0 & 19.4 ± 3.39 & 557.09 ± 98.48 \\
\bottomrule
\end{tabular}
\end{adjustbox}
\caption{Results for artificial univariate data with error scale $3$, $n=30,000$}
\label{tab:placeholder}
\end{table}
\clearpage

\subsection{Multivariate data results} \label{app:multivariateresults}

\begin{table}[H]
\begin{adjustbox}{max width=\linewidth, max height=11cm}
\begin{tabular}{lllllllll}
\toprule
 &  &  &  &  &  & crps & mse & time \\
dataset & data size & error type & error scale & model size & predictive &  &  &  \\
\midrule
\multirow[t]{119}{*}{Multivariate complex} & \multirow[t]{119}{*}{10000} & \multirow[t]{23}{*}{gamma} & \multirow[t]{11}{*}{3} & \multirow[t]{2}{*}{2x128} & gaussian & 0.738 ± 0.015 & 5.638 ± 0.224 & 317.043 ± 53.49 \\
 &  &  &  &  & student t & 0.699 ± 0.005*** & 5.148 ± 0.078*** & 308.574 ± 35.597 \\
\cline{5-9}
 &  &  &  & \multirow[t]{3}{*}{2x64} & gaussian & 0.75 ± 0.016 & 5.82 ± 0.225 & 313.059 ± 67.642 \\
 &  &  &  &  & skew & 0.755 ± 0.012 & 5.893 ± 0.181 & 368.657 ± 56.568 \\
 &  &  &  &  & student t & 0.7 ± 0.008*** & 5.195 ± 0.118*** & 330.529 ± 42.949 \\
\cline{5-9}
 &  &  &  & \multirow[t]{3}{*}{4x128} & gaussian & 0.773 ± 0.007 & 5.979 ± 0.116 & 693.709 ± 78.698 \\
 &  &  &  &  & skew & 0.796 ± 0.009*** & 6.333 ± 0.123*** & 508.75 ± 44.073 \\
 &  &  &  &  & student t & 0.731 ± 0.006*** & 5.43 ± 0.06*** & 531.712 ± 37.099 \\
\cline{5-9}
 &  &  &  & \multirow[t]{3}{*}{4x64} & gaussian & 1.289 ± 0.01 & 25.996 ± 0.204 & 112.459 ± 30.806 \\
 &  &  &  &  & skew & 1.218 ± 0.161* & 23.354 ± 6.402* & 213.312 ± 143.672 \\
 &  &  &  &  & student t & 1.054 ± 0.245*** & 17.43 ± 9.641*** & 285.316 ± 181.998 \\
\cline{4-9} \cline{5-9}
 &  &  & \multirow[t]{12}{*}{15} & \multirow[t]{3}{*}{2x128} & gaussian & 2.773 ± 0.089 & 80.366 ± 6.108 & 195.608 ± 59.325 \\
 &  &  &  &  & skew & 2.728 ± 0.008* & 77.352 ± 0.299* & 291.905 ± 31.829 \\
 &  &  &  &  & student t & 2.663 ± 0.003*** & 76.717 ± 0.338** & 192.352 ± 19.128 \\
\cline{5-9}
 &  &  &  & \multirow[t]{3}{*}{2x64} & gaussian & 2.799 ± 0.12 & 82.675 ± 7.995 & 198.616 ± 79.565 \\
 &  &  &  &  & skew & 2.756 ± 0.087 & 79.572 ± 5.753 & 299.638 ± 74.35 \\
 &  &  &  &  & student t & 2.657 ± 0.004*** & 76.71 ± 0.263*** & 186.287 ± 16.039 \\
\cline{5-9}
 &  &  &  & \multirow[t]{3}{*}{4x128} & gaussian & 3.003 ± 0.01 & 96.409 ± 0.502 & 190.815 ± 40.872 \\
 &  &  &  &  & skew & 2.996 ± 0.011* & 95.707 ± 0.42*** & 160.704 ± 36.752 \\
 &  &  &  &  & student t & 2.725 ± 0.014*** & 78.566 ± 0.387*** & 303.491 ± 32.472 \\
\cline{5-9}
 &  &  &  & \multirow[t]{3}{*}{4x64} & gaussian & 3.004 ± 0.009 & 96.935 ± 0.332 & 125.811 ± 13.799 \\
 &  &  &  &  & skew & 3.008 ± 0.011 & 96.961 ± 0.37 & 136.781 ± 22.101 \\
 &  &  &  &  & student t & 2.889 ± 0.114*** & 91.171 ± 7.751*** & 189.276 ± 92.018 \\
\cline{3-9} \cline{4-9} \cline{5-9}
 &  & \multirow[t]{24}{*}{gaussian} & \multirow[t]{12}{*}{3} & \multirow[t]{3}{*}{2x128} & gaussian & 0.731 ± 0.009 & 5.512 ± 0.117 & 345.973 ± 54.926 \\
 &  &  &  &  & skew & 0.732 ± 0.01 & 5.539 ± 0.138 & 387.758 ± 54.571 \\
 &  &  &  &  & student t & 0.704 ± 0.007*** & 5.143 ± 0.098*** & 390.4 ± 38.192 \\
\cline{5-9}
 &  &  &  & \multirow[t]{3}{*}{2x64} & gaussian & 0.75 ± 0.012 & 5.805 ± 0.174 & 304.125 ± 37.716 \\
 &  &  &  &  & skew & 0.746 ± 0.013 & 5.735 ± 0.178 & 365.964 ± 63.68 \\
 &  &  &  &  & student t & 0.713 ± 0.007*** & 5.271 ± 0.091*** & 342.752 ± 44.896 \\
\cline{5-9}
 &  &  &  & \multirow[t]{3}{*}{4x128} & gaussian & 0.761 ± 0.012 & 5.787 ± 0.148 & 536.072 ± 66.842 \\
 &  &  &  &  & skew & 0.789 ± 0.006*** & 6.187 ± 0.097*** & 606.006 ± 45.453 \\
 &  &  &  &  & student t & 0.739 ± 0.007*** & 5.454 ± 0.101*** & 544.811 ± 54.905 \\
\cline{5-9}
 &  &  &  & \multirow[t]{3}{*}{4x64} & gaussian & 1.289 ± 0.015 & 25.672 ± 0.33 & 149.528 ± 62.303 \\
 &  &  &  &  & skew & 1.278 ± 0.019* & 25.403 ± 0.496* & 178.29 ± 67.112 \\
 &  &  &  &  & student t & 0.902 ± 0.23*** & 11.407 ± 8.811*** & 439.859 ± 201.148 \\
\cline{4-9} \cline{5-9}
 &  &  & \multirow[t]{12}{*}{15} & \multirow[t]{3}{*}{2x128} & gaussian & 2.782 ± 0.005 & 76.697 ± 0.31 & 212.24 ± 17.448 \\
 &  &  &  &  & skew & 2.784 ± 0.004 & 76.822 ± 0.259 & 237.832 ± 16.006 \\
 &  &  &  &  & student t & 2.786 ± 0.007** & 76.715 ± 0.408 & 171.581 ± 14.175 \\
\cline{5-9}
 &  &  &  & \multirow[t]{3}{*}{2x64} & gaussian & 2.813 ± 0.078 & 78.937 ± 5.109 & 210.636 ± 51.736 \\
 &  &  &  &  & skew & 2.786 ± 0.007 & 77.162 ± 0.423 & 284.501 ± 24.11 \\
 &  &  &  &  & student t & 2.778 ± 0.005* & 76.494 ± 0.277* & 196.199 ± 16.879 \\
\cline{5-9}
 &  &  &  & \multirow[t]{3}{*}{4x128} & gaussian & 3.056 ± 0.01 & 94.637 ± 0.805 & 140.764 ± 32.641 \\
 &  &  &  &  & skew & 3.051 ± 0.014 & 94.143 ± 0.898* & 159.024 ± 49.548 \\
 &  &  &  &  & student t & 2.915 ± 0.096*** & 82.874 ± 7.503*** & 277.985 ± 98.854 \\
\cline{5-9}
 &  &  &  & \multirow[t]{3}{*}{4x64} & gaussian & 3.056 ± 0.006 & 95.25 ± 0.257 & 119.42 ± 30.112 \\
 &  &  &  &  & skew & 3.059 ± 0.006* & 95.478 ± 0.493* & 132.325 ± 27.666 \\
 &  &  &  &  & student t & 3.056 ± 0.006 & 94.661 ± 0.477*** & 131.154 ± 30.143 \\
\cline{3-9} \cline{4-9} \cline{5-9}
 &  & \multirow[t]{24}{*}{laplace} & \multirow[t]{12}{*}{3} & \multirow[t]{3}{*}{2x128} & gaussian & 0.736 ± 0.007 & 5.679 ± 0.091 & 292.645 ± 40.975 \\
 &  &  &  &  & skew & 0.729 ± 0.009** & 5.58 ± 0.144** & 402.194 ± 56.431 \\
 &  &  &  &  & student t & 0.681 ± 0.006*** & 4.985 ± 0.068*** & 373.918 ± 36.259 \\
\cline{5-9}
 &  &  &  & \multirow[t]{3}{*}{2x64} & gaussian & 0.755 ± 0.004 & 5.957 ± 0.073 & 276.499 ± 29.997 \\
 &  &  &  &  & skew & 0.751 ± 0.005*** & 5.89 ± 0.072*** & 385.87 ± 44.791 \\
 &  &  &  &  & student t & 0.723 ± 0.018*** & 5.544 ± 0.249*** & 301.747 ± 63.02 \\
\cline{5-9}
 &  &  &  & \multirow[t]{3}{*}{4x128} & gaussian & 0.814 ± 0.131 & 7.621 ± 5.317 & 465.819 ± 100.779 \\
 &  &  &  &  & skew & 0.793 ± 0.013 & 6.335 ± 0.218 & 626.21 ± 61.535 \\
 &  &  &  &  & student t & 0.757 ± 0.018* & 5.712 ± 0.21 & 465.364 ± 56.783 \\
\cline{5-9}
 &  &  &  & \multirow[t]{3}{*}{4x64} & gaussian & 1.294 ± 0.007 & 26.995 ± 0.094 & 126.962 ± 27.822 \\
 &  &  &  &  & skew & 1.29 ± 0.011 & 26.852 ± 0.187*** & 177.239 ± 46.521 \\
 &  &  &  &  & student t & 1.271 ± 0.008*** & 26.712 ± 0.19*** & 155.808 ± 36.23 \\
\cline{4-9} \cline{5-9}
 &  &  & \multirow[t]{12}{*}{15} & \multirow[t]{3}{*}{2x128} & gaussian & 2.707 ± 0.091 & 77.962 ± 5.477 & 208.003 ± 51.153 \\
 &  &  &  &  & skew & 2.688 ± 0.077 & 76.839 ± 4.821 & 288.746 ± 58.561 \\
 &  &  &  &  & student t & 2.612 ± 0.005*** & 73.997 ± 0.261*** & 216.888 ± 20.279 \\
\cline{5-9}
 &  &  &  & \multirow[t]{3}{*}{2x64} & gaussian & 2.915 ± 0.008 & 91.593 ± 0.51 & 99.779 ± 25.604 \\
 &  &  &  &  & skew & 2.919 ± 0.01 & 91.75 ± 0.506 & 114.812 ± 28.657 \\
 &  &  &  &  & student t & 2.825 ± 0.105*** & 87.38 ± 6.545** & 145.656 ± 66.907 \\
\cline{5-9}
 &  &  &  & \multirow[t]{3}{*}{4x128} & gaussian & 2.953 ± 0.008 & 93.125 ± 0.51 & 142.873 ± 16.094 \\
 &  &  &  &  & skew & 2.952 ± 0.012 & 93.04 ± 0.447 & 149.473 ± 24.076 \\
 &  &  &  &  & student t & 2.932 ± 0.011*** & 93.301 ± 0.47 & 131.823 ± 14.741 \\
\cline{5-9}
 &  &  &  & \multirow[t]{3}{*}{4x64} & gaussian & 2.938 ± 0.008 & 93.009 ± 0.305 & 127.979 ± 19.289 \\
 &  &  &  &  & skew & 2.938 ± 0.008 & 93.006 ± 0.375 & 145.273 ± 18.779 \\
 &  &  &  &  & student t & 2.903 ± 0.007*** & 92.417 ± 0.497*** & 160.327 ± 26.987 \\
\bottomrule
\end{tabular}
\end{adjustbox}
\caption{Results for multivariate artificial dataset, $n=10,000$. - part 1}
\end{table}

\begin{table}[H]
\centering
\begin{adjustbox}{max width=\linewidth, max height=12cm}
\begin{tabular}{lllllllll}
\toprule
 &  &  &  &  &  & crps & mse & time \\
dataset & data size & error type & error scale & model size & predictive &  &  &  \\
\midrule
\multirow[t]{119}{*}{Multivariate complex} & \multirow[t]{119}{*}{10000} & \multirow[t]{24}{*}{lognormal} & \multirow[t]{12}{*}{3} & \multirow[t]{3}{*}{2x128} & gaussian & 1.296 ± 0.004 & 27.562 ± 0.081 & 66.27 ± 14.202 \\
 &  &  &  &  & skew & 1.297 ± 0.006 & 27.562 ± 0.105 & 90.861 ± 19.424 \\
 &  &  &  &  & student t & 1.291 ± 0.004*** & 28.095 ± 0.147*** & 62.29 ± 9.704 \\
\cline{5-9}
 &  &  &  & \multirow[t]{3}{*}{2x64} & gaussian & 1.292 ± 0.002 & 27.541 ± 0.065 & 76.457 ± 12.122 \\
 &  &  &  &  & skew & 1.292 ± 0.002 & 27.526 ± 0.063 & 107.081 ± 13.977 \\
 &  &  &  &  & student t & 1.288 ± 0.003*** & 27.99 ± 0.163*** & 86.686 ± 14.819 \\
\cline{5-9}
 &  &  &  & \multirow[t]{3}{*}{4x128} & gaussian & 1.31 ± 0.026 & 27.711 ± 0.11 & 97.514 ± 16.342 \\
 &  &  &  &  & skew & 1.317 ± 0.039 & 27.757 ± 0.22 & 125.196 ± 17.5 \\
 &  &  &  &  & student t & 1.309 ± 0.024 & 28.197 ± 0.187*** & 106.483 ± 11.447 \\
\cline{5-9}
 &  &  &  & \multirow[t]{3}{*}{4x64} & gaussian & 1.296 ± 0.005 & 27.648 ± 0.078 & 119.156 ± 18.937 \\
 &  &  &  &  & skew & 1.295 ± 0.004 & 27.656 ± 0.075 & 142.989 ± 23.22 \\
 &  &  &  &  & student t & 1.29 ± 0.004*** & 28.078 ± 0.106*** & 112.74 ± 15.323 \\
\cline{4-9} \cline{5-9}
 &  &  & \multirow[t]{12}{*}{15} & \multirow[t]{3}{*}{2x128} & gaussian & 2.728 ± 0.014 & 92.965 ± 0.334 & 70.197 ± 7.96 \\
 &  &  &  &  & skew & 2.736 ± 0.021 & 93.051 ± 0.335 & 88.929 ± 11.019 \\
 &  &  &  &  & student t & 2.563 ± 0.011*** & 94.826 ± 0.438*** & 73.209 ± 8.222 \\
\cline{5-9}
 &  &  &  & \multirow[t]{3}{*}{2x64} & gaussian & 2.715 ± 0.009 & 92.911 ± 0.358 & 78.35 ± 12.247 \\
 &  &  &  &  & skew & 2.714 ± 0.01 & 92.887 ± 0.281 & 98.482 ± 13.684 \\
 &  &  &  &  & student t & 2.539 ± 0.005*** & 94.675 ± 0.477*** & 78.624 ± 9.697 \\
\cline{5-9}
 &  &  &  & \multirow[t]{3}{*}{4x128} & gaussian & 2.828 ± 0.086 & 94.053 ± 1.055 & 108.975 ± 10.962 \\
 &  &  &  &  & skew & 2.85 ± 0.109 & 94.167 ± 1.323 & 126.627 ± 10.171 \\
 &  &  &  &  & student t & 2.688 ± 0.091*** & 96.471 ± 1.295*** & 112.975 ± 11.979 \\
\cline{5-9}
 &  &  &  & \multirow[t]{3}{*}{4x64} & gaussian & 2.721 ± 0.02 & 93.123 ± 0.319 & 118.911 ± 20.939 \\
 &  &  &  &  & skew & 2.716 ± 0.013 & 93.005 ± 0.238 & 137.597 ± 18.623 \\
 &  &  &  &  & student t & 2.557 ± 0.018*** & 95.141 ± 0.537*** & 131.984 ± 19.508 \\
\cline{3-9} \cline{4-9} \cline{5-9}
 &  & \multirow[t]{24}{*}{student t} & \multirow[t]{12}{*}{3} & \multirow[t]{3}{*}{2x128} & gaussian & 1.016 ± 0.236 & 16.486 ± 9.472 & 178.464 ± 82.844 \\
 &  &  &  &  & skew & 1.039 ± 0.234 & 17.022 ± 9.193 & 260.515 ± 132.516 \\
 &  &  &  &  & student t & 1.226 ± 0.092*** & 25.864 ± 3.957*** & 134.269 ± 34.559 \\
\cline{5-9}
 &  &  &  & \multirow[t]{3}{*}{2x64} & gaussian & 1.308 ± 0.003 & 27.84 ± 0.089 & 80.489 ± 11.742 \\
 &  &  &  &  & skew & 1.306 ± 0.002*** & 27.752 ± 0.117** & 106.194 ± 15.238 \\
 &  &  &  &  & student t & 1.287 ± 0.012*** & 27.776 ± 0.249 & 96.825 ± 13.431 \\
\cline{5-9}
 &  &  &  & \multirow[t]{3}{*}{4x128} & gaussian & 1.313 ± 0.007 & 27.537 ± 0.22 & 159.454 ± 14.037 \\
 &  &  &  &  & skew & 1.313 ± 0.01 & 27.526 ± 0.208 & 177.415 ± 21.3 \\
 &  &  &  &  & student t & 1.363 ± 0.041*** & 28.347 ± 0.34*** & 149.82 ± 13.837 \\
\cline{5-9}
 &  &  &  & \multirow[t]{3}{*}{4x64} & gaussian & 1.309 ± 0.002 & 27.778 ± 0.121 & 126.63 ± 15.176 \\
 &  &  &  &  & skew & 1.308 ± 0.003 & 27.736 ± 0.073 & 163.63 ± 19.805 \\
 &  &  &  &  & student t & 1.3 ± 0.011*** & 27.878 ± 0.102** & 158.734 ± 25.554 \\
\cline{4-9} \cline{5-9}
 &  &  & \multirow[t]{12}{*}{15} & \multirow[t]{3}{*}{2x128} & gaussian & 2.853 ± 0.008 & 99.23 ± 0.437 & 89.027 ± 13.289 \\
 &  &  &  &  & skew & 2.861 ± 0.01*** & 99.39 ± 0.57 & 95.472 ± 12.569 \\
 &  &  &  &  & student t & 2.788 ± 0.01*** & 98.992 ± 0.586 & 88.217 ± 11.299 \\
\cline{5-9}
 &  &  &  & \multirow[t]{3}{*}{2x64} & gaussian & 2.842 ± 0.009 & 98.991 ± 0.432 & 87.204 ± 12.208 \\
 &  &  &  &  & skew & 2.845 ± 0.008 & 99.069 ± 0.383 & 115.067 ± 17.45 \\
 &  &  &  &  & student t & 2.773 ± 0.01*** & 98.67 ± 0.505* & 102.463 ± 13.483 \\
\cline{5-9}
 &  &  &  & \multirow[t]{3}{*}{4x128} & gaussian & 2.913 ± 0.029 & 100.135 ± 0.556 & 141.723 ± 15.905 \\
 &  &  &  &  & skew & 2.925 ± 0.032 & 100.471 ± 0.773 & 158.492 ± 17.325 \\
 &  &  &  &  & student t & 3.124 ± 0.112*** & 104.601 ± 2.655*** & 124.9 ± 8.478 \\
\cline{5-9}
 &  &  &  & \multirow[t]{3}{*}{4x64} & gaussian & 2.855 ± 0.015 & 99.451 ± 0.706 & 141.511 ± 16.661 \\
 &  &  &  &  & skew & 2.851 ± 0.011 & 99.202 ± 0.408 & 182.882 ± 19.162 \\
 &  &  &  &  & student t & 2.812 ± 0.029*** & 99.417 ± 0.77 & 152.887 ± 17.985 \\
\bottomrule
\end{tabular}
\end{adjustbox}
\caption{Results for multivariate artificial dataset, $n=10,000$. - part 2}
\label{tab:placeholder}
\end{table}

\begin{table}[H]
\centering
\begin{adjustbox}{max width=\linewidth, max height=11cm}
\begin{tabular}{lllllllll}
\toprule
 &  &  &  &  &  & crps & mse & time \\
dataset & data size & error type & error scale & model size & predictive &  &  &  \\
\midrule
\multirow[t]{72}{*}{Multivariate complex} & \multirow[t]{72}{*}{50000} & \multirow[t]{24}{*}{gamma} & \multirow[t]{12}{*}{3} & \multirow[t]{3}{*}{2x128} & gaussian & 0.686 ± 0.002 & 5.036 ± 0.024 & 1050.153 ± 148.485 \\
 &  &  &  &  & skew & 0.69 ± 0.002*** & 5.073 ± 0.047*** & 1158.334 ± 99.48 \\
 &  &  &  &  & student t & 0.664 ± 0.007*** & 4.83 ± 0.079*** & 998.522 ± 165.715 \\
\cline{5-9}
 &  &  &  & \multirow[t]{3}{*}{2x64} & gaussian & 0.687 ± 0.002 & 5.055 ± 0.033 & 1044.029 ± 123.477 \\
 &  &  &  &  & skew & 0.685 ± 0.002** & 5.037 ± 0.034* & 1424.431 ± 172.171 \\
 &  &  &  &  & student t & 0.673 ± 0.004*** & 4.94 ± 0.041*** & 1001.564 ± 140.909 \\
\cline{5-9}
 &  &  &  & \multirow[t]{3}{*}{4x128} & gaussian & 0.699 ± 0.005 & 5.204 ± 0.14 & 1491.709 ± 256.88 \\
 &  &  &  &  & skew & 0.713 ± 0.014*** & 5.335 ± 0.186** & 1907.71 ± 297.492 \\
 &  &  &  &  & student t & 0.693 ± 0.01** & 5.15 ± 0.134 & 1372.274 ± 130.293 \\
\cline{5-9}
 &  &  &  & \multirow[t]{3}{*}{4x64} & gaussian & 0.709 ± 0.008 & 5.298 ± 0.076 & 1581.01 ± 316.187 \\
 &  &  &  &  & skew & 0.717 ± 0.008*** & 5.429 ± 0.156*** & 1797.844 ± 331.422 \\
 &  &  &  &  & student t & 0.697 ± 0.015*** & 5.195 ± 0.194** & 1508.355 ± 462.915 \\
\cline{4-9} \cline{5-9}
 &  &  & \multirow[t]{12}{*}{15} & \multirow[t]{3}{*}{2x128} & gaussian & 2.693 ± 0.01 & 76.155 ± 0.585 & 713.112 ± 131.509 \\
 &  &  &  &  & skew & 2.693 ± 0.014 & 76.042 ± 0.415 & 860.081 ± 166.378 \\
 &  &  &  &  & student t & 2.655 ± 0.009*** & 77.135 ± 0.943*** & 520.787 ± 133.338 \\
\cline{5-9}
 &  &  &  & \multirow[t]{3}{*}{2x64} & gaussian & 2.677 ± 0.013 & 75.491 ± 0.306 & 691.012 ± 157.026 \\
 &  &  &  &  & skew & 2.687 ± 0.013** & 76.038 ± 0.55*** & 712.903 ± 175.638 \\
 &  &  &  &  & student t & 2.648 ± 0.008*** & 76.65 ± 0.751*** & 568.155 ± 84.472 \\
\cline{5-9}
 &  &  &  & \multirow[t]{3}{*}{4x128} & gaussian & 2.722 ± 0.024 & 77.097 ± 0.732 & 1005.287 ± 187.261 \\
 &  &  &  &  & skew & 2.733 ± 0.028 & 77.498 ± 1.125 & 1016.23 ± 236.927 \\
 &  &  &  &  & student t & 2.681 ± 0.016*** & 77.08 ± 0.984 & 943.549 ± 170.919 \\
\cline{5-9}
 &  &  &  & \multirow[t]{3}{*}{4x64} & gaussian & 2.704 ± 0.013 & 76.744 ± 0.418 & 1317.221 ± 226.578 \\
 &  &  &  &  & skew & 2.708 ± 0.01 & 76.766 ± 0.468 & 1581.615 ± 317.723 \\
 &  &  &  &  & student t & 2.656 ± 0.007*** & 76.791 ± 0.565 & 996.099 ± 265.448 \\
\cline{3-9} \cline{4-9} \cline{5-9}
 &  & \multirow[t]{24}{*}{gaussian} & \multirow[t]{12}{*}{3} & \multirow[t]{3}{*}{2x128} & gaussian & 0.68 ± 0.004 & 4.88 ± 0.044 & 1099.229 ± 127.503 \\
 &  &  &  &  & skew & 0.682 ± 0.006 & 4.907 ± 0.064 & 1354.78 ± 156.765 \\
 &  &  &  &  & student t & 0.671 ± 0.003*** & 4.734 ± 0.041*** & 1006.065 ± 168.248 \\
\cline{5-9}
 &  &  &  & \multirow[t]{3}{*}{2x64} & gaussian & 0.696 ± 0.003 & 5.091 ± 0.038 & 1138.438 ± 147.482 \\
 &  &  &  &  & skew & 0.698 ± 0.002* & 5.108 ± 0.029 & 1434.663 ± 194.798 \\
 &  &  &  &  & student t & 0.677 ± 0.005*** & 4.823 ± 0.055*** & 926.754 ± 99.859 \\
\cline{5-9}
 &  &  &  & \multirow[t]{3}{*}{4x128} & gaussian & 0.708 ± 0.002 & 5.224 ± 0.038 & 1239.774 ± 131.021 \\
 &  &  &  &  & skew & 0.717 ± 0.014** & 5.322 ± 0.213* & 1957.411 ± 270.889 \\
 &  &  &  &  & student t & 0.704 ± 0.003*** & 5.097 ± 0.04*** & 1438.853 ± 420.988 \\
\cline{5-9}
 &  &  &  & \multirow[t]{3}{*}{4x64} & gaussian & 0.703 ± 0.003 & 5.182 ± 0.054 & 2034.157 ± 614.587 \\
 &  &  &  &  & skew & 0.71 ± 0.009*** & 5.286 ± 0.146*** & 1911.359 ± 224.631 \\
 &  &  &  &  & student t & 0.696 ± 0.013** & 5.044 ± 0.172*** & 1681.976 ± 361.0 \\
\cline{4-9} \cline{5-9}
 &  &  & \multirow[t]{12}{*}{15} & \multirow[t]{3}{*}{2x128} & gaussian & 2.741 ± 0.007 & 74.616 ± 0.308 & 555.871 ± 164.751 \\
 &  &  &  &  & skew & 2.747 ± 0.007*** & 75.053 ± 0.408*** & 816.8 ± 143.163 \\
 &  &  &  &  & student t & 2.747 ± 0.006*** & 74.751 ± 0.376 & 560.918 ± 82.643 \\
\cline{5-9}
 &  &  &  & \multirow[t]{3}{*}{2x64} & gaussian & 2.745 ± 0.004 & 75.054 ± 0.314 & 670.145 ± 107.255 \\
 &  &  &  &  & skew & 2.743 ± 0.006 & 75.122 ± 0.356 & 901.531 ± 193.99 \\
 &  &  &  &  & student t & 2.745 ± 0.006 & 74.787 ± 0.306** & 683.956 ± 140.162 \\
\cline{5-9}
 &  &  &  & \multirow[t]{3}{*}{4x128} & gaussian & 2.768 ± 0.005 & 75.397 ± 0.2 & 927.574 ± 74.199 \\
 &  &  &  &  & skew & 2.781 ± 0.007*** & 76.206 ± 0.256*** & 967.233 ± 57.794 \\
 &  &  &  &  & student t & 2.772 ± 0.004** & 75.208 ± 0.281** & 811.762 ± 95.306 \\
\cline{5-9}
 &  &  &  & \multirow[t]{3}{*}{4x64} & gaussian & 2.757 ± 0.004 & 75.347 ± 0.175 & 1143.245 ± 97.06 \\
 &  &  &  &  & skew & 2.758 ± 0.005 & 75.547 ± 0.258*** & 1243.762 ± 128.473 \\
 &  &  &  &  & student t & 2.763 ± 0.006*** & 75.41 ± 0.33 & 1065.736 ± 225.402 \\
\cline{3-9} \cline{4-9} \cline{5-9}
 &  & \multirow[t]{24}{*}{laplace} & \multirow[t]{12}{*}{3} & \multirow[t]{3}{*}{2x128} & gaussian & 0.677 ± 0.011 & 4.989 ± 0.144 & 1061.945 ± 237.739 \\
 &  &  &  &  & skew & 0.683 ± 0.01* & 5.062 ± 0.126* & 1405.713 ± 358.729 \\
 &  &  &  &  & student t & 0.642 ± 0.005*** & 4.583 ± 0.073*** & 1115.436 ± 294.805 \\
\cline{5-9}
 &  &  &  & \multirow[t]{3}{*}{2x64} & gaussian & 0.697 ± 0.016 & 5.247 ± 0.233 & 1193.405 ± 241.145 \\
 &  &  &  &  & skew & 0.687 ± 0.005** & 5.134 ± 0.065* & 1501.251 ± 270.549 \\
 &  &  &  &  & student t & 0.659 ± 0.006*** & 4.787 ± 0.064*** & 1061.533 ± 122.6 \\
\cline{5-9}
 &  &  &  & \multirow[t]{3}{*}{4x128} & gaussian & 0.704 ± 0.012 & 5.289 ± 0.166 & 1507.463 ± 320.404 \\
 &  &  &  &  & skew & 0.702 ± 0.009 & 5.339 ± 0.246 & 1703.275 ± 393.687 \\
 &  &  &  &  & student t & 0.688 ± 0.006*** & 5.075 ± 0.063*** & 1301.759 ± 306.285 \\
\cline{5-9}
 &  &  &  & \multirow[t]{3}{*}{4x64} & gaussian & 0.707 ± 0.012 & 5.369 ± 0.152 & 1600.415 ± 330.729 \\
 &  &  &  &  & skew & 0.708 ± 0.013 & 5.377 ± 0.156 & 1983.979 ± 366.771 \\
 &  &  &  &  & student t & 0.691 ± 0.01*** & 5.243 ± 0.182** & 1473.606 ± 257.391 \\
\cline{4-9} \cline{5-9}
 &  &  & \multirow[t]{12}{*}{15} & \multirow[t]{3}{*}{2x128} & gaussian & 2.657 ± 0.012 & 76.239 ± 0.682 & 758.778 ± 219.271 \\
 &  &  &  &  & skew & 2.66 ± 0.007 & 76.077 ± 0.29 & 902.774 ± 181.711 \\
 &  &  &  &  & student t & 2.625 ± 0.009*** & 75.527 ± 0.458*** & 641.507 ± 214.587 \\
\cline{5-9}
 &  &  &  & \multirow[t]{3}{*}{2x64} & gaussian & 2.682 ± 0.065 & 77.863 ± 4.478 & 783.522 ± 79.552 \\
 &  &  &  &  & skew & 2.674 ± 0.032 & 77.211 ± 1.734 & 1179.525 ± 350.979 \\
 &  &  &  &  & student t & 2.623 ± 0.004*** & 75.645 ± 0.144* & 665.061 ± 163.258 \\
\cline{5-9}
 &  &  &  & \multirow[t]{3}{*}{4x128} & gaussian & 2.805 ± 0.147 & 82.504 ± 9.034 & 1214.956 ± 469.133 \\
 &  &  &  &  & skew & 2.922 ± 0.228* & 86.396 ± 10.765 & 1040.694 ± 303.593 \\
 &  &  &  &  & student t & 2.692 ± 0.021*** & 77.15 ± 0.647** & 916.148 ± 156.736 \\
\cline{5-9}
 &  &  &  & \multirow[t]{3}{*}{4x64} & gaussian & 2.735 ± 0.075 & 79.922 ± 6.007 & 1388.094 ± 260.478 \\
 &  &  &  &  & skew & 2.704 ± 0.022* & 77.913 ± 1.378 & 1754.169 ± 236.173 \\
 &  &  &  &  & student t & 2.66 ± 0.01*** & 76.212 ± 0.432** & 1282.184 ± 370.811 \\
\bottomrule
\end{tabular}
\end{adjustbox}
\caption{Results for multivariate artificial dataset, $n=50,000$. - part 1}
\label{tab:placeholder}
\end{table}

\begin{table}[H]
\centering
\begin{adjustbox}{max width=\linewidth, max height=12cm}
\begin{tabular}{lllllllll}
\toprule
 &  &  &  &  &  & crps & mse & time \\
dataset & data size & error type & error scale & model size & predictive &  &  &  \\
\midrule
\multirow[t]{47}{*}{Multivariate complex} & \multirow[t]{47}{*}{50000} & \multirow[t]{23}{*}{lognormal} & \multirow[t]{11}{*}{3} & \multirow[t]{3}{*}{2x128} & gaussian & 1.373 ± 0.032 & 27.89 ± 0.203 & 258.37 ± 33.772 \\
 &  &  &  &  & skew & 1.353 ± 0.042* & 27.717 ± 0.206** & 307.169 ± 40.543 \\
 &  &  &  &  & student t & 1.343 ± 0.022*** & 28.135 ± 0.308*** & 253.845 ± 22.899 \\
\cline{5-9}
 &  &  &  & \multirow[t]{3}{*}{2x64} & gaussian & 1.314 ± 0.019 & 27.591 ± 0.094 & 289.493 ± 82.468 \\
 &  &  &  &  & skew & 1.304 ± 0.023 & 27.442 ± 0.403 & 404.52 ± 98.542 \\
 &  &  &  &  & student t & 1.278 ± 0.069** & 26.952 ± 2.508 & 381.832 ± 265.098 \\
\cline{5-9}
 &  &  &  & \multirow[t]{2}{*}{4x128} & gaussian & 2.053 ± 0.322 & 46.91 ± 10.962 & 338.48 ± 40.535 \\
 &  &  &  &  & student t & 1.851 ± 0.41* & 41.876 ± 14.96 & 426.484 ± 91.815 \\
\cline{5-9}
 &  &  &  & \multirow[t]{3}{*}{4x64} & gaussian & 1.474 ± 0.149 & 30.172 ± 3.181 & 382.365 ± 62.614 \\
 &  &  &  &  & skew & 1.609 ± 0.165** & 32.028 ± 3.263* & 505.764 ± 85.385 \\
 &  &  &  &  & student t & 1.478 ± 0.115 & 30.5 ± 2.293 & 463.363 ± 80.606 \\
\cline{4-9} \cline{5-9}
 &  &  & \multirow[t]{12}{*}{15} & \multirow[t]{3}{*}{2x128} & gaussian & 3.002 ± 0.126 & 95.598 ± 3.097 & 265.088 ± 29.456 \\
 &  &  &  &  & skew & 2.944 ± 0.079* & 93.906 ± 2.126* & 312.617 ± 32.106 \\
 &  &  &  &  & student t & 2.83 ± 0.138*** & 96.937 ± 2.82 & 276.397 ± 12.852 \\
\cline{5-9}
 &  &  &  & \multirow[t]{3}{*}{2x64} & gaussian & 2.731 ± 0.037 & 90.487 ± 0.673 & 268.175 ± 27.983 \\
 &  &  &  &  & skew & 2.758 ± 0.097 & 91.225 ± 2.064 & 328.702 ± 43.938 \\
 &  &  &  &  & student t & 2.544 ± 0.021*** & 91.896 ± 0.72*** & 297.788 ± 45.65 \\
\cline{5-9}
 &  &  &  & \multirow[t]{3}{*}{4x128} & gaussian & 5.971 ± 2.022 & 438.25 ± 361.872 & 406.613 ± 66.428 \\
 &  &  &  &  & skew & 5.592 ± 1.003 & 345.636 ± 204.008 & 459.592 ± 61.041 \\
 &  &  &  &  & student t & 5.251 ± 1.265 & 361.316 ± 227.659 & 427.134 ± 61.935 \\
\cline{5-9}
 &  &  &  & \multirow[t]{3}{*}{4x64} & gaussian & 3.455 ± 0.515 & 135.294 ± 57.637 & 454.449 ± 82.672 \\
 &  &  &  &  & skew & 3.131 ± 0.167** & 106.468 ± 9.576* & 475.161 ± 80.912 \\
 &  &  &  &  & student t & 3.036 ± 0.402*** & 115.788 ± 28.763 & 474.725 ± 65.479 \\
\cline{3-9} \cline{4-9} \cline{5-9}
 &  & \multirow[t]{24}{*}{student t} & \multirow[t]{12}{*}{3} & \multirow[t]{3}{*}{2x128} & gaussian & 0.745 ± 0.027 & 6.025 ± 0.43 & 820.213 ± 346.016 \\
 &  &  &  &  & skew & 0.748 ± 0.026 & 6.102 ± 0.417 & 930.198 ± 307.77 \\
 &  &  &  &  & student t & 0.765 ± 0.14 & 7.8 ± 6.019 & 900.987 ± 193.42 \\
\cline{5-9}
 &  &  &  & \multirow[t]{3}{*}{2x64} & gaussian & 1.16 ± 0.203 & 22.015 ± 9.192 & 456.826 ± 213.735 \\
 &  &  &  &  & skew & 1.103 ± 0.224 & 19.169 ± 9.958 & 879.777 ± 393.139 \\
 &  &  &  &  & student t & 1.005 ± 0.238** & 16.579 ± 10.419* & 756.485 ± 314.1 \\
\cline{5-9}
 &  &  &  & \multirow[t]{3}{*}{4x128} & gaussian & 1.161 ± 0.222 & 21.782 ± 9.214 & 813.476 ± 506.225 \\
 &  &  &  &  & skew & 1.246 ± 0.478 & 24.317 ± 19.179 & 1087.71 ± 711.319 \\
 &  &  &  &  & student t & 1.116 ± 0.418 & 18.615 ± 12.717 & 853.219 ± 413.609 \\
\cline{5-9}
 &  &  &  & \multirow[t]{3}{*}{4x64} & gaussian & 1.453 ± 0.18 & 29.929 ± 3.276 & 484.868 ± 94.796 \\
 &  &  &  &  & skew & 1.369 ± 0.256 & 26.376 ± 8.364* & 681.041 ± 338.031 \\
 &  &  &  &  & student t & 0.943 ± 0.257*** & 14.563 ± 10.348*** & 1227.529 ± 565.605 \\
\cline{4-9} \cline{5-9}
 &  &  & \multirow[t]{12}{*}{15} & \multirow[t]{3}{*}{2x128} & gaussian & 2.764 ± 0.12 & 88.102 ± 9.001 & 548.05 ± 289.483 \\
 &  &  &  &  & skew & 2.839 ± 0.152* & 91.574 ± 7.748 & 483.938 ± 209.898 \\
 &  &  &  &  & student t & 2.639 ± 0.149*** & 83.66 ± 9.431 & 708.78 ± 203.399 \\
\cline{5-9}
 &  &  &  & \multirow[t]{3}{*}{2x64} & gaussian & 2.83 ± 0.018 & 93.989 ± 1.117 & 347.724 ± 65.022 \\
 &  &  &  &  & skew & 2.844 ± 0.017** & 94.307 ± 1.151 & 513.919 ± 140.419 \\
 &  &  &  &  & student t & 2.642 ± 0.095*** & 84.128 ± 6.507*** & 793.557 ± 152.352 \\
\cline{5-9}
 &  &  &  & \multirow[t]{3}{*}{4x128} & gaussian & 4.931 ± 2.471 & 338.877 ± 340.788 & 372.521 ± 22.363 \\
 &  &  &  &  & skew & 5.641 ± 2.08 & 378.124 ± 334.243 & 553.122 ± 96.927 \\
 &  &  &  &  & student t & 4.038 ± 0.843 & 162.548 ± 60.412** & 400.386 ± 39.102 \\
\cline{5-9}
 &  &  &  & \multirow[t]{3}{*}{4x64} & gaussian & 3.459 ± 0.531 & 131.676 ± 56.35 & 479.382 ± 36.735 \\
 &  &  &  &  & skew & 3.505 ± 0.835 & 136.96 ± 83.667 & 550.354 ± 75.216 \\
 &  &  &  &  & student t & 2.831 ± 0.263*** & 91.43 ± 11.351*** & 758.315 ± 326.225 \\
\cline{1-9} \cline{2-9} \cline{3-9} \cline{4-9} \cline{5-9}
\bottomrule
\end{tabular}

\end{adjustbox}
\caption{Results for multivariate artificial dataset, $n=50,000$. - part 2}
\label{tab:placeholder}
\end{table}
\clearpage

\subsection{Real world data results} \label{app:realworldresults}

\begin{table}[H]
\centering
\begin{adjustbox}{max width=\linewidth, max height=12cm}
\begin{tabular}{lllllllll}
\toprule
 &  &  & crps & epoch & mae & mape & mse & time \\
dataset & model size & likelihood distribution &  &  &  &  &  &  \\
\midrule
\multirow[t]{27}{*}{ENTSOE} & \multirow[t]{3}{*}{1×16} & gaussian & 2579.95 ± 356.2 & 64.53 ± 8.76 & 3658.4 ± 560.45 & 6.8 ± 1.11 & 22722824.0 ± 6271888.5 & 33.61 ± 4.56 \\
 &  & skew & 2638.57 ± 363.98 & 63.83 ± 10.94 & 3763.2 ± 562.25 & 6.98 ± 1.13 & 23792096.0 ± 6299335.5 & 43.04 ± 7.36 \\
 &  & student's $t$ & 2160.71 ± 161.02 & 67.37 ± 6.11 & 2976.08 ± 233.51 & 5.5 ± 0.44 & 16120954.0 ± 2108091.75 & 36.38 ± 3.3 \\
\cline{2-9}
 & \multirow[t]{3}{*}{1×32} & gaussian & 1973.2 ± 170.03 & 71.53 ± 6.47 & 2690.78 ± 272.88 & 4.94 ± 0.53 & 13178415.0 ± 2474580.75 & 35.8 ± 3.23 \\
 &  & skew & 2066.07 ± 202.88 & 65.5 ± 7.37 & 2832.49 ± 324.61 & 5.22 ± 0.63 & 14396662.0 ± 2878910.5 & 44.58 ± 4.99 \\
 &  & student's $t$ & 1686.17 ± 63.57 & 65.23 ± 7.35 & 2248.8 ± 101.45 & 4.13 ± 0.18 & 9934459.0 ± 722850.81 & 35.06 ± 4.01 \\
\cline{2-9}
 & \multirow[t]{3}{*}{1×4} & gaussian & 3321.78 ± 345.85 & 74.57 ± 18.63 & 4808.25 ± 422.83 & 9.0 ± 0.88 & 36327744.0 ± 6857106.0 & 36.53 ± 9.08 \\
 &  & skew & 3815.16 ± 2.94 & 45.3 ± 4.42 & 5408.88 ± 5.81 & 10.25 ± 0.01 & 46161040.0 ± 64043.61 & 30.68 ± 2.98 \\
 &  & student's $t$ & 3040.81 ± 20.23 & 81.33 ± 6.04 & 4220.85 ± 25.65 & 8.05 ± 0.05 & 31441000.0 ± 222416.17 & 43.62 ± 3.25 \\
\cline{2-9}
 & \multirow[t]{3}{*}{4×16} & gaussian & 3154.04 ± 175.46 & 66.33 ± 11.88 & 4579.87 ± 226.09 & 8.55 ± 0.46 & 32555590.0 ± 3585925.0 & 66.97 ± 11.94 \\
 &  & skew & 3390.01 ± 316.79 & 58.4 ± 15.78 & 4866.36 ± 406.2 & 9.15 ± 0.83 & 37060836.0 ± 6655838.0 & 69.79 ± 18.81 \\
 &  & student's $t$ & 3079.86 ± 37.81 & 62.4 ± 5.17 & 4234.36 ± 63.75 & 8.08 ± 0.13 & 31438682.0 ± 358155.84 & 66.3 ± 5.47 \\
\cline{2-9}
 & \multirow[t]{3}{*}{4×32} & gaussian & 3076.05 ± 17.67 & 52.37 ± 4.29 & 4428.9 ± 33.46 & 8.31 ± 0.05 & 31112602.0 ± 258982.84 & 53.94 ± 4.33 \\
 &  & skew & 3060.07 ± 32.93 & 53.7 ± 7.21 & 4400.73 ± 48.56 & 8.25 ± 0.1 & 30828756.0 ± 405906.03 & 65.12 ± 8.71 \\
 &  & student's $t$ & 3036.03 ± 39.67 & 46.4 ± 6.08 & 4144.28 ± 60.42 & 7.87 ± 0.13 & 30791126.0 ± 400261.19 & 49.09 ± 6.55 \\
\cline{2-9}
 & \multirow[t]{3}{*}{4×4} & gaussian & 3816.2 ± 1.84 & 43.07 ± 2.71 & 5409.75 ± 3.77 & 10.26 ± 0.01 & 46197652.0 ± 50350.48 & 44.71 ± 2.82 \\
 &  & skew & 3816.78 ± 1.82 & 43.93 ± 3.21 & 5410.66 ± 3.38 & 10.26 ± 0.01 & 46198680.0 ± 48954.91 & 52.72 ± 3.85 \\
 &  & student's $t$ & 3844.21 ± 3.08 & 40.7 ± 3.11 & 5271.68 ± 3.38 & 10.14 ± 0.01 & 47362188.0 ± 89780.24 & 43.23 ± 3.29 \\
\cline{2-9}
 & \multirow[t]{3}{*}{8×16} & gaussian & 3774.48 ± 64.08 & 39.73 ± 9.19 & 5355.33 ± 69.93 & 10.15 ± 0.15 & 44950124.0 ± 1529085.13 & 69.13 ± 15.96 \\
 &  & skew & 3813.75 ± 6.42 & 36.57 ± 3.16 & 5407.95 ± 9.48 & 10.25 ± 0.02 & 46079460.0 ± 186373.88 & 69.48 ± 5.95 \\
 &  & student's $t$ & 3309.29 ± 195.98 & 72.83 ± 15.12 & 4553.81 ± 267.78 & 8.7 ± 0.5 & 34984708.0 ± 3845178.25 & 123.74 ± 25.8 \\
\cline{2-9}
 & \multirow[t]{3}{*}{8×32} & gaussian & 3147.28 ± 17.57 & 60.27 ± 5.72 & 4535.06 ± 52.24 & 8.51 ± 0.05 & 31890740.0 ± 276827.25 & 100.98 ± 9.51 \\
 &  & skew & 3150.85 ± 35.11 & 68.53 ± 8.82 & 4582.99 ± 102.12 & 8.53 ± 0.11 & 31964494.0 ± 563581.81 & 127.43 ± 16.23 \\
 &  & student's $t$ & 3132.67 ± 22.96 & 55.23 ± 5.19 & 4319.99 ± 25.65 & 8.23 ± 0.07 & 31747636.0 ± 367234.72 & 97.28 ± 9.15 \\
\cline{2-9}
 & \multirow[t]{3}{*}{8×4} & gaussian & 3816.97 ± 1.76 & 40.9 ± 2.72 & 5412.08 ± 3.78 & 10.26 ± 0.01 & 46177544.0 ± 49957.5 & 67.72 ± 4.69 \\
 &  & skew & 3817.39 ± 1.98 & 42.47 ± 3.11 & 5411.44 ± 3.34 & 10.26 ± 0.01 & 46178600.0 ± 47883.08 & 80.15 ± 5.96 \\
 &  & student's $t$ & 3842.87 ± 2.72 & 39.2 ± 2.63 & 5272.75 ± 3.87 & 10.14 ± 0.01 & 47324400.0 ± 70216.66 & 69.63 ± 4.64 \\
\cline{1-9} \cline{2-9}
\bottomrule
\end{tabular}
\end{adjustbox}
\caption{Results for ENTSO-E dataset.}
\label{tab:placeholder}
\end{table}

\begin{table}[H]
\centering
\begin{adjustbox}{max width=\linewidth, max height=12cm}
\begin{tabular}{lllllllll}
\toprule
 &  &  & crps & epoch & mae & mape & mse & time \\
dataset & model size & likelihood distribution &  &  &  &  &  &  \\
\midrule
\multirow[t]{26}{*}{POWER} & \multirow[t]{3}{*}{1×16} & gaussian & 1.79 ± 0.01 & 50.17 ± 3.69 & 2.54 ± 0.02 & 4.99 ± 0.06 & 10.85 ± 0.11 & 143.31 ± 10.53 \\
 &  & skew & 1.78 ± 0.01 & 52.13 ± 4.17 & 2.54 ± 0.02 & 4.97 ± 0.05 & 10.83 ± 0.12 & 188.18 ± 14.62 \\
 &  & student's $t$ & 1.77 ± 0.01 & 44.03 ± 3.85 & 2.51 ± 0.01 & 4.91 ± 0.04 & 10.77 ± 0.1 & 126.12 ± 11.01 \\
\cline{2-9}
 & \multirow[t]{3}{*}{1×32} & gaussian & 1.78 ± 0.01 & 48.73 ± 3.79 & 2.53 ± 0.01 & 4.97 ± 0.05 & 10.8 ± 0.08 & 138.3 ± 11.05 \\
 &  & skew & 1.78 ± 0.01 & 49.43 ± 3.69 & 2.53 ± 0.01 & 4.95 ± 0.05 & 10.74 ± 0.06 & 174.66 ± 12.99 \\
 &  & student's $t$ & 1.77 ± 0.01 & 44.07 ± 3.41 & 2.5 ± 0.01 & 4.88 ± 0.05 & 10.72 ± 0.08 & 124.56 ± 9.43 \\
\cline{2-9}
 & \multirow[t]{3}{*}{1×4} & gaussian & 1.82 ± 0.02 & 49.6 ± 3.83 & 2.58 ± 0.02 & 5.0 ± 0.06 & 11.16 ± 0.2 & 139.51 ± 10.83 \\
 &  & skew & 1.82 ± 0.01 & 56.3 ± 5.86 & 2.58 ± 0.02 & 5.01 ± 0.05 & 11.2 ± 0.16 & 204.33 ± 21.36 \\
 &  & student's $t$ & 1.79 ± 0.02 & 45.93 ± 3.68 & 2.54 ± 0.02 & 4.94 ± 0.05 & 10.98 ± 0.19 & 134.45 ± 10.67 \\
\cline{2-9}
 & \multirow[t]{3}{*}{4×16} & gaussian & 1.79 ± 0.01 & 60.6 ± 5.46 & 2.52 ± 0.02 & 4.93 ± 0.06 & 10.86 ± 0.14 & 337.42 ± 31.76 \\
 &  & skew & 1.79 ± 0.01 & 68.77 ± 3.86 & 2.53 ± 0.02 & 4.94 ± 0.07 & 10.88 ± 0.17 & 435.52 ± 24.48 \\
 &  & student's $t$ & 1.78 ± 0.01 & 50.67 ± 3.63 & 2.51 ± 0.02 & 4.93 ± 0.07 & 10.81 ± 0.16 & 277.14 ± 19.73 \\
\cline{2-9}
 & \multirow[t]{3}{*}{4×32} & gaussian & 1.8 ± 0.01 & 52.8 ± 3.64 & 2.52 ± 0.02 & 4.97 ± 0.07 & 10.84 ± 0.18 & 284.61 ± 18.95 \\
 &  & skew & 1.8 ± 0.01 & 59.93 ± 4.56 & 2.52 ± 0.02 & 4.95 ± 0.06 & 10.91 ± 0.17 & 380.76 ± 29.54 \\
 &  & student's $t$ & 1.78 ± 0.01 & 46.57 ± 4.42 & 2.5 ± 0.02 & 4.89 ± 0.06 & 10.78 ± 0.16 & 256.05 ± 23.86 \\
\cline{2-9}
 & \multirow[t]{3}{*}{4×4} & gaussian & 7.38 ± 0.02 & 20.67 ± 2.83 & 10.99 ± 0.04 & 21.47 ± 0.19 & 176.05 ± 0.86 & 112.44 ± 15.06 \\
 &  & skew & 7.38 ± 0.02 & 21.77 ± 2.7 & 10.99 ± 0.04 & 21.46 ± 0.18 & 176.15 ± 0.9 & 133.16 ± 16.32 \\
 &  & student's $t$ & 4.25 ± 2.77 & 45.0 ± 20.77 & 6.2 ± 4.13 & 11.89 ± 7.82 & 82.56 ± 81.42 & 252.43 ± 116.25 \\
\cline{2-9}
 & \multirow[t]{3}{*}{8×16} & gaussian & 7.19 ± 0.99 & 23.83 ± 12.04 & 10.69 ± 1.5 & 20.87 ± 2.92 & 169.78 ± 29.4 & 211.51 ± 105.97 \\
 &  & skew & 6.46 ± 2.05 & 35.27 ± 34.23 & 9.59 ± 3.11 & 18.74 ± 6.09 & 148.19 ± 61.01 & 339.41 ± 329.5 \\
 &  & student's $t$ & 2.21 ± 1.36 & 66.47 ± 14.26 & 3.11 ± 2.02 & 6.01 ± 3.74 & 21.81 ± 38.69 & 604.8 ± 131.81 \\
\cline{2-9}
 & \multirow[t]{2}{*}{8×32} & gaussian & 1.92 ± 0.04 & 71.37 ± 4.69 & 2.59 ± 0.03 & 5.08 ± 0.08 & 11.66 ± 0.34 & 650.0 ± 41.72 \\
  &  & skew & 1.90 ± 0.02 & 97.33 ± 3.66 & 2.59 ± 0.02 & 5.03 ± 0.06 & 11.53 ± 0.31 & 981.07 ± 31.25 \\
 &  & student's $t$ & 1.89 ± 0.04 & 58.47 ± 4.51 & 2.55 ± 0.02 & 4.97 ± 0.08 & 11.35 ± 0.33 & 528.05 ± 44.08 \\
\cline{2-9}
 & \multirow[t]{3}{*}{8×4} & gaussian & 7.39 ± 0.02 & 19.6 ± 2.44 & 11.01 ± 0.02 & 21.55 ± 0.12 & 176.33 ± 0.68 & 176.75 ± 22.01 \\
 &  & skew & 7.39 ± 0.01 & 20.43 ± 2.81 & 11.01 ± 0.02 & 21.56 ± 0.11 & 176.34 ± 0.69 & 197.5 ± 27.05 \\
 &  & student's $t$ & 7.43 ± 0.01 & 19.3 ± 2.35 & 10.96 ± 0.03 & 21.02 ± 0.19 & 176.97 ± 0.62 & 178.91 ± 21.62 \\
\cline{1-9} \cline{2-9}
\bottomrule
\end{tabular}
\end{adjustbox}
\caption{Results for POWER dataset.}
\label{tab:placeholder}
\end{table}

\section*{Acknowledgments}
This work is supported by the Helmholtz Association Initiative and Networking Fund under the Helmholtz AI platform grant. 

\bibliographystyle{siamplain}
\bibliography{references}

@article{knoblauch2019generalized,
  title={Generalized variational inference: Three arguments for deriving new posteriors},
  author={Knoblauch, Jeremias and Jewson, Jack and Damoulas, Theodoros},
  journal={arXiv preprint arXiv:1904.02063},
  year={2019}
}

@inproceedings{shah2014student,
  title={Student-t processes as alternatives to Gaussian processes},
  author={Shah, Amar and Wilson, Andrew and Ghahramani, Zoubin},
  booktitle={Artificial intelligence and statistics},
  pages={877--885},
  year={2014},
  organization={PMLR}
}

@article{jylanki2011robust,
  title={Robust Gaussian Process Regression with a Student-t Likelihood.},
  author={Jyl{\"a}nki, Pasi and Vanhatalo, Jarno and Vehtari, Aki},
  journal={Journal of Machine Learning Research},
  volume={12},
  number={11},
  year={2011}
}

@article{huggins2019robust,
  title={Robust inference and model criticism using bagged posteriors},
  author={Huggins, Jonathan H and Miller, Jeffrey W},
  journal={arXiv preprint arXiv:1912.07104},
  year={2019}
}

@article{gelman2017prior,
  title={The prior can often only be understood in the context of the likelihood},
  author={Gelman, Andrew and Simpson, Daniel and Betancourt, Michael},
  journal={Entropy},
  volume={19},
  number={10},
  pages={555},
  year={2017},
  publisher={MDPI}
}

@article{fortuin2021bayesian,
  title={Bayesian neural network priors revisited},
  author={Fortuin, Vincent and Garriga-Alonso, Adri{\`a} and Ober, Sebastian W and Wenzel, Florian and R{\"a}tsch, Gunnar and Turner, Richard E and van der Wilk, Mark and Aitchison, Laurence},
  journal={arXiv preprint arXiv:2102.06571},
  year={2021}
}

@inproceedings{dusenberry2020efficient,
  title={Efficient and scalable bayesian neural nets with rank-1 factors},
  author={Dusenberry, Michael and Jerfel, Ghassen and Wen, Yeming and Ma, Yian and Snoek, Jasper and Heller, Katherine and Lakshminarayanan, Balaji and Tran, Dustin},
  booktitle={International conference on machine learning},
  pages={2782--2792},
  year={2020},
  organization={PMLR}
}

@article{blei2017variational,
  title={Variational inference: A review for statisticians},
  author={Blei, David M and Kucukelbir, Alp and McAuliffe, Jon D},
  journal={Journal of the American statistical Association},
  volume={112},
  number={518},
  pages={859--877},
  year={2017},
  publisher={Taylor \& Francis}
}

@article{hoffman2013stochastic,
  title={Stochastic variational inference},
  author={Hoffman, Matthew D and Blei, David M and Wang, Chong and Paisley, John},
  journal={the Journal of machine Learning research},
  volume={14},
  number={1},
  pages={1303--1347},
  year={2013},
  publisher={JMLR. org}
}

@inproceedings{blundell2015weight,
  title={Weight uncertainty in neural network},
  author={Blundell, Charles and Cornebise, Julien and Kavukcuoglu, Koray and Wierstra, Daan},
  booktitle={International conference on machine learning},
  pages={1613--1622},
  year={2015},
  organization={PMLR}
}

@inproceedings{khan2018fast,
  title={Fast and scalable bayesian deep learning by weight-perturbation in adam},
  author={Khan, Mohammad and Nielsen, Didrik and Tangkaratt, Voot and Lin, Wu and Gal, Yarin and Srivastava, Akash},
  booktitle={International conference on machine learning},
  pages={2611--2620},
  year={2018},
  organization={PMLR}
}

@article{kingma2015variational,
  title={Variational dropout and the local reparameterization trick},
  author={Kingma, Durk P and Salimans, Tim and Welling, Max},
  journal={Advances in neural information processing systems},
  volume={28},
  year={2015}
}

@article{kullback1951information,
  title={On information and sufficiency},
  author={Kullback, Solomon and Leibler, Richard A},
  journal={The annals of mathematical statistics},
  volume={22},
  number={1},
  pages={79--86},
  year={1951},
  publisher={JSTOR}
}

@inproceedings{izmailov2021bayesian,
  title={What are Bayesian neural network posteriors really like?},
  author={Izmailov, Pavel and Vikram, Sharad and Hoffman, Matthew D and Wilson, Andrew Gordon Gordon},
  booktitle={International conference on machine learning},
  pages={4629--4640},
  year={2021},
  organization={PMLR}
}

@techreport{neal1992bayesian,
  title={Bayesian training of backpropagation networks by the hybrid Monte Carlo method},
  author={Neal, Radford M},
  year={1992},
  institution={Technical Report CRG-TR-92-1, Dept. of Computer Science, University of Toronto}
}

@article{mackay1995bayesian,
  title={Bayesian neural networks and density networks},
  author={MacKay, David JC},
  journal={Nuclear Instruments and Methods in Physics Research Section A: Accelerators, Spectrometers, Detectors and Associated Equipment},
  volume={354},
  number={1},
  pages={73--80},
  year={1995},
  publisher={Elsevier}
}

@inproceedings{vladimirova2019understanding,
  title={Understanding priors in Bayesian neural networks at the unit level},
  author={Vladimirova, Mariia and Verbeek, Jakob and Mesejo, Pablo and Arbel, Julyan},
  booktitle={International Conference on Machine Learning},
  pages={6458--6467},
  year={2019},
  organization={PMLR}
}

@article{noci2021precise,
  title={Precise characterization of the prior predictive distribution of deep ReLU networks},
  author={Noci, Lorenzo and Bachmann, Gregor and Roth, Kevin and Nowozin, Sebastian and Hofmann, Thomas},
  journal={Advances in Neural Information Processing Systems},
  volume={34},
  pages={20851--20862},
  year={2021}
}

@article{gneiting2007strictly,
  title={Strictly proper scoring rules, prediction, and estimation},
  author={Gneiting, Tilmann and Raftery, Adrian E},
  journal={Journal of the American statistical Association},
  volume={102},
  number={477},
  pages={359--378},
  year={2007},
  publisher={Taylor \& Francis}
}

@misc{combined_cycle_power_plant_294,
  author       = {Tfekci, Pnar and Kaya, Heysem},
  title        = {{Combined Cycle Power Plant}},
  year         = {2014},
  howpublished = {UCI Machine Learning Repository},
  note         = {{DOI}: https://doi.org/10.24432/C5002N}
}

@article{platform2022entso,
  title={ENTSO-E transparency platform},
  author={Platform, ENTSO-E Transparency},
  journal={Avalable: https://transparency. entsoe. eu/dashboard/show},
  year={2022}
}

@inproceedings{kinga2015method,
  title={A method for stochastic optimization},
  author={Kinga, Diederik and Adam, Jimmy Ba and others},
  booktitle={International conference on learning representations (ICLR)},
  volume={5},
  number={6},
  year={2015},
  organization={California;}
}

@article{kendall2017uncertainties,
  title={What uncertainties do we need in bayesian deep learning for computer vision?},
  author={Kendall, Alex and Gal, Yarin},
  journal={Advances in neural information processing systems},
  volume={30},
  year={2017}
}

@article{hong2016probabilistic,
  title={Probabilistic electric load forecasting: A tutorial review},
  author={Hong, Tao and Fan, Shu},
  journal={International Journal of Forecasting},
  volume={32},
  number={3},
  pages={914--938},
  year={2016},
  publisher={Elsevier}
}

@article{armstrong2023deep,
  title={A deep-learning phase picker with calibrated Bayesian-derived uncertainties for earthquakes in the Yellowstone volcanic region},
  author={Armstrong, Alysha D and Claerhout, Zachary and Baker, Ben and Koper, Keith D},
  journal={Bulletin of the Seismological Society of America},
  volume={113},
  number={6},
  pages={2323--2344},
  year={2023},
  publisher={Seismological Society of America}
}

@article{nagl2022quantifying,
  title={Quantifying uncertainty of machine learning methods for loss given default},
  author={Nagl, Matthias and Nagl, Maximilian and R{\"o}sch, Daniel},
  journal={Frontiers in Applied Mathematics and Statistics},
  volume={8},
  pages={1076083},
  year={2022},
  publisher={Frontiers Media SA}
}

@misc{EUAIAct2024,
  title        = {Regulation (EU) 2024/1689 of the European Parliament and of the Council of 13 June 2024 laying down harmonised rules on artificial intelligence (Artificial Intelligence Act)},
  howpublished = {Official Journal of the European Union, OJ L, 12 July 2024},
  year         = {2024},
  url          = {https://eur-lex.europa.eu/eli/reg/2024/1689/oj},
  note         = {See Article 15: Accuracy, robustness and cybersecurity}
}

@inproceedings{ansel2024pytorch,
  title={Pytorch 2: Faster machine learning through dynamic python bytecode transformation and graph compilation},
  author={Ansel, Jason and Yang, Edward and He, Horace and Gimelshein, Natalia and Jain, Animesh and Voznesensky, Michael and Bao, Bin and Bell, Peter and Berard, David and Burovski, Evgeni and others},
  booktitle={Proceedings of the 29th ACM International Conference on Architectural Support for Programming Languages and Operating Systems, Volume 2},
  pages={929--947},
  year={2024}
}

@inproceedings{hernandez-lobato_probabilistic_2015,
	location = {Lille, France},
	title = {Probabilistic backpropagation for scalable learning of Bayesian neural networks},
	series = {{ICML}'15},
    year={2015},
	pages = {1861--1869},
	booktitle = {Proceedings of the 32nd International Conference on International Conference on Machine Learning - Volume 37},
	publisher = {{JMLR}.org},
	author = {Hernández-Lobato, José Miguel and Adams, Ryan P.},
	urldate = {2024-07-23},
	date = {2015-07-06},
}

@article{ghosh2019model,
  title={Model selection in Bayesian neural networks via horseshoe priors},
  author={Ghosh, Soumya and Yao, Jiayu and Doshi-Velez, Finale},
  journal={Journal of Machine Learning Research},
  volume={20},
  number={182},
  pages={1--46},
  year={2019}
}

@inproceedings{carvalho_handling_nodate,
  title={Handling sparsity via the horseshoe},
  author={Carvalho, Carlos M and Polson, Nicholas G and Scott, James G},
  booktitle={Artificial intelligence and statistics},
  pages={73--80},
  year={2009},
  organization={PMLR}
}

@article{zhang2018VI_advances,
  title={Advances in variational inference},
  author={Zhang, Cheng and B{\"u}tepage, Judith and Kjellstr{\"o}m, Hedvig and Mandt, Stephan},
  journal={IEEE transactions on pattern analysis and machine intelligence},
  volume={41},
  number={8},
  pages={2008--2026},
  year={2018},
  publisher={IEEE}
}

@incollection{neal1998view,
  title={A view of the EM algorithm that justifies incremental, sparse, and other variants},
  author={Neal, Radford M and Hinton, Geoffrey E},
  booktitle={Learning in graphical models},
  pages={355--368},
  year={1998},
  publisher={Springer}
}

@misc{torchbayesian,
  title = {torch bayesian - Easy Variational Inference},
  howpublished = {\url{https://github.com/RAI-SCC/torch_bayesian/releases/tag/v0.0.1}}
}

@article{briegel2000dynamic,
  title={Dynamic neural regression models},
  author={Briegel, Thomas and Tresp, Volker},
  year={2000}
}

@article{lee2021scale,
  title={Scale mixtures of neural network Gaussian processes},
  author={Lee, Hyungi and Yun, Eunggu and Yang, Hongseok and Lee, Juho},
  journal={arXiv preprint arXiv:2107.01408},
  year={2021}
}

@article{saad2024scalable,
  title={Scalable spatiotemporal prediction with Bayesian neural fields},
  author={Saad, Feras and Burnim, Jacob and Carroll, Colin and Patton, Brian and K{\"o}ster, Urs and A. Saurous, Rif and Hoffman, Matthew},
  journal={Nature Communications},
  volume={15},
  number={1},
  pages={7942},
  year={2024},
  publisher={Nature Publishing Group UK London}
}

@article{lampinen2001bayesian,
  title={Bayesian approach for neural networks—review and case studies},
  author={Lampinen, Jouko and Vehtari, Aki},
  journal={Neural networks},
  volume={14},
  number={3},
  pages={257--274},
  year={2001},
  publisher={Elsevier}
}

@inproceedings{futami2018variational,
  title={Variational inference based on robust divergences},
  author={Futami, Futoshi and Sato, Issei and Sugiyama, Masashi},
  booktitle={International Conference on Artificial Intelligence and Statistics},
  pages={813--822},
  year={2018},
  organization={PMLR}
}
\end{document}